\documentclass[11pt]{article}

\usepackage{geometry}
\usepackage{setspace}
\usepackage{caption} 
\usepackage{subcaption}
\usepackage{graphicx}

\usepackage{mathptmx}
\usepackage[T1]{fontenc}

\usepackage{threeparttable}

\usepackage[obeyFinal]{todonotes}
\presetkeys{todonotes}{backgroundcolor=yellow}{}
\usepackage[inline,shortlabels]{enumitem}
\usepackage{multirow}
\usepackage[hidelinks]{hyperref}
\usepackage[]{xcolor}
\usepackage{wrapfig}
\usepackage{lscape}
\usepackage{rotating}
\usepackage{xspace}
\usepackage[section]{placeins}
\usepackage{booktabs}
\usepackage{colortbl}
\usepackage{hhline}
\usepackage{multicol}
\usepackage{multirow}
\usepackage{adjustbox}
\usepackage{tabularx}

\usepackage{amsmath}

\newcommand{\elements}{process model elements\xspace}

\newcommand{\diagram}{process model diagram\xspace}

\newcommand{\diagrams}{process model diagrams\xspace}

\newcommand{\processdescription}{process model description\xspace}

\newcommand{\yes}{Yes} %\textcolor{darkgreen}{Yes}}
\newcommand{\no}{No} %\textcolor{red}{No}}
\newcommand{\sota}{state-of-the-art\xspace}

\newcommand{\pet}{process model extraction from text\xspace}
\newcommand{\Pet}{Process model extraction from text\xspace}
\newcommand{\PET}{Process Model Extraction from Text\xspace}
\newcommand{\PETD}{PET dataset\xspace}
\newcommand{\petonly}{PET\xspace}

\newcommand{\fa}{f_{a}\xspace}
\newcommand{\fb}{f_{b}\xspace}
\newcommand{\papers}{10\xspace} %%%%%% AT THE END - FIX THIS WITH THE CORRECT NUMBER!
\newcommand{\tools}{6\xspace} %%%%%% AT THE END - FIX THIS WITH THE CORRECT NUMBER!
\newcommand{\declare}{DECLARE\xspace}
\newcommand{\gitrepo}{\url{github.com/patriziobellan86/ProcessExtractionFromTextSotaAndChallenges}\xspace}
\newcommand{\mypar}[1]{\vspace{0.5pt}\noindent\textbf{#1.}}

\usepackage[natbibapa]{apacite}

\newenvironment{biseabstract}{%
\begin{quote} \bf}
{\end{quote}}

\newenvironment{bisekeywords}{%
\begin{quote} \it \textbf{Keywords -}}
{\end{quote}}

\title{Process Extraction from Text: Benchmarking the State of the Art and Paving the Way for Future Challenges}
\author
{Patrizio Bellan$^{1,2\ast}$, Mauro Dragoni$^{1}$, Chiara Ghidini$^{1}$, Han van der Aa$^{3}$, Simone Ponzetto$^{3}$\\
\\
\normalsize{$^{1}$Fondazione Bruno Kessler, Povo (Tn), Italy}\\
\normalsize{$^{2}$Free University of Bozen-Bolzano, Bolzano, Italy}\\
\normalsize{$^{3}$University of Mannheim, Mannheim, Germany}\\
\\
\normalsize{$^\ast$To whom correspondence should be addressed; E-mail:  pbellan@fbk.eu}
}
\date{}

\begin{document} 

\baselineskip24pt


\maketitle

% Place your abstract within the special {biseabstract} environment.
\begin{biseabstract}
  %!TEX root = .././main.tex

The extraction of process models from text refers to the problem of turning the information contained in an unstructured textual process descriptions into a formal representation, i.e., a process model. 
Several automated approaches have been proposed to tackle this problem, but they are highly heterogeneous in scope (what they exactly do) and underlying assumptions, such as differences in input, target output, and data used in their evaluation. As a result, it is currently unclear how well existing solutions are able to solve the model-extraction problem and how they compare to each other.
In this paper, we overcome this issue by comparing \papers state-of-the-art approaches for model extraction in a systematic manner, covering both qualitative and quantitative aspects. 

The qualitative evaluation compares the analysis of the primary studies on:
\begin{enumerate*}[(i)]
	\item the main characteristics of each solution; 
	\item the type of \elements extracted from the input data; and	
	\item the experimental evaluation performed to evaluate the proposed framework.
\end{enumerate*} The results show a heterogeneity of techniques, elements extracted and evaluations conducted, that are often impossible to compare in a direct manner. To overcome this difficulty we propose a quantitative comparison of the tools proposed by the different papers on the unifying task of process model entity and relation extraction so as to be able to compare them directly. The results show three distinct groups of tools in terms of performance, with no tool obtaining very good scores and also serious limitations in terms of elements extracted. Moreover, the proposed evaluation pipeline can be considered a reference task on a well defined dataset and metrics that can be used to compare new tools in a way similar to what happens in e.g., the natural language processing or computer vision communities. 
The paper also presents a reflection on the results of the qualitative and quantitative evaluation on the limitations and challenges that the community needs to address in the future to produce significant advances in this area.

%\textit{\Pet} can be regarded as the specific problem of finding an algorithmic function to transform textual process descriptions, of varied and different nature, into their formal representation (the process model diagram). While being an active and important field of research, this area still lacks a systematic comparative analysis of the work and the tools produced so far that can provide a comprehensive picture and an articulated baseline for future works in this area. 
%%
%In this paper we analyze, in a comparative manner, reference state-of-the-art literature and available tools, both from a qualitative and quantitative perspective, especially for what concerns the techniques used, the process elements extracted and the evaluations performed. As a result of the analysis, we discuss important limitations that hamper the exploitation of recent Natural Language Processing techniques in this field and we discuss fundamental limitations and challenges for the future concerning the datasets, the techniques, the experimental evaluations, and the pipelines currently adopted and to be developed in the future. 
%%

\end{biseabstract}

\begin{bisekeywords}
Process Model Extraction from Text, 
Process Discovery,
Business Process Management,
Process Model,
Natural Language Processing
\end{bisekeywords}
%
%\begin{biseacknowledgements}
%I'd like to say thanks to... Please add the acknowledgements \textbf{only in the final version after acceptance} in case they would identify or hint towards the authors.
%\end{biseacknowledgements}
%
%
%%%%%%%%%%

 %!TEX root = .././main.tex

\section{Introduction}
\label{sec:introduction}

\textit{\Pet} can be regarded as the specific problem of finding an algorithmic function to transform textual process descriptions into their formal representation (the process model diagram).
The ambiguous nature of natural language, the multiple possible writing styles, and the great variability of possible domains of application make this task extremely challenging. Indeed recent papers on this topic~\citep{Riefer16,Maqbool18, Aa18} highlight that after almost ten years of research, \pet~is a task far from being resolved and further research in this direction is needed to improve the quality of the process model generation. 

One of the issues of \pet is the lack of systematic efforts to tackle this problem. Indeed, works in this area often address self-defined, specific sub-tasks of (e.g., the extraction of specific sets of entities / relations occurring in a process model), using different data, with different techniques and evaluation procedures. This has made the comparison of individual works difficult, and has hampered the incremental production of better and better techniques to address the overall \pet challenge.   

In this paper, we tackle the fragmentation of this research field by proposing qualitative and quantitative comparative analyses of well-known recent works, to better understand their contributions and limitations and to identify key challenges for the research community to address. To scope our study, we have focused on works published in the area of Business Process Management since 2010,  which target both declarative (e.g., \declare) and procedural (e.g., BPMN) languages. 
%The specific methodology adopted for the choice of papers, and their content, are described in Section~\ref{sec:protocol}. The methodology relies upon papers already selected in previous surveys, complemented by recent relevant papers published at the two flagship conferences for research on business process analysis, i.e., the  International Conference on Advanced Information Systems Engineering (CAiSE) and the International Conference on Business Process Management Conference (BPM).
Differently from recent surveys, we do not limit ourselves to the provision of a literature review.
The novelty of this paper is threefold: first, we provide a detailed \emph{qualitative} comparison of the selected contributions in terms of approaches and techniques used, extracted process elements and conducted experimental evaluations (Section~\ref{sec:qualitative-comparison}); second we provide a first ever comparative \emph{quantitative} analysis of the contribution for which a tool was available using the recently published PET dataset~\citep{DBLP:conf/bpm/BellanADGP22} (Section~\ref{sec:quantitative-comparison}); third, we present a discussion of both the qualitative and quantitative comparisons (Section~\ref{sec:discussion}), which highlights \emph{open challenges} for the research community to address. 

The analysis performed on the \papers papers identifies four groups of problems: (i) problems with the limited and rather specific data used for training and testing, which fail to cover the variety of process description styles existing in the real world; (ii) problems with the techniques adopted, which often fail to incorporate novel NLP techniques, and thus produce tools unable to provide satisfactory and scalable results; (iii) the lack of a common evaluation framework where to comparatively judge the quality of the proposed systems, not only in terms of solid data benchmarks but also in terms of evaluation metrics able to measure a good extracted process model (or some of its characteristics); and finally (iv) problems related to the entire (complex) pipeline that has to be constructed to provide \pet solutions. 
While the systematic analysis contained in the paper (and especially the \emph{quantitative} analysis) can be provided a first step towards the construction of a common evaluation framework for the \pet field, and indeed provide a first reference comparative benchmarking analysis of the different approaches available in literature, the limitations identified in Section~\ref{sec:discussion} and the questions reported for each group of problems can provide the basis for a discussion on relevant challenges that the community needs to solve in order to make the task of \pet scale to real-world scenarios.

 \section{Related Work}
\label{sec:related-works}

The importance of \pet has been widely recognized in literature, being included among the main opportunities that NLP can provide in the context of BPM. We can roughly divide the papers that try to give an overview of this field in two sets. The first one contains papers that are mainly focused on challenges and opportunities that pertain process modeling in general, including \pet and the intersection of NLP techniques and \pet. Significant examples of these papers are ~\citep{DBLP:journals/dss/MendlingBBF17,Mendling:2014aa,Aa18,DBLP:conf/refsq/MendlingLTA19}. The second set contains instead on papers which compare previous works in \pet for different purposes. Reference works in this set are \cite{Riefer16, Maqbool18, Bordignon2018NaturalLP, Indahyanti2022AutoGeneratingBP}. Our work pertains mainly to the second set and in this Section we provide a careful comparison with the existing work within this set, which is also summarized in \autoref{tab:rwoverview}. In fact the main purpose of our work is to benchmark the state of the art with a thorough qualitative and quantitative evaluation with a final discussion of limitations and challenges for the field (Section~\ref{sec:discussion}) which stems from the comparative analysis  performed. 

\begin{table}[!htbp]
	\caption{Related work on the analysis of text-to-model transformation approaches.}
	\label{tab:rwoverview}
	\small
	\begin{tabularx}{\textwidth}{p{2.4cm}p{3cm}p{2cm}X}
		\toprule
		\textbf{Paper} & \textbf{Focus} & \textbf{Coverage} & \textbf{Performed analyses} \\
		\midrule
	\cite{Riefer16} & Text-to-model transformation approaches & 5 papers (2007--2015) & Qualitatively compares input, method, and output of approaches. \\
	
	\cite{Maqbool18} & Text-to-BPMN transformation approaches and associated articles & 36 papers (2011--2018) & Categorizes NLP techniques, tools, and BPMN elements covered by papers that propose text-to-BPMN approaches or associated techniques. \\
	\cite{Bordignon2018NaturalLP} & Approaches using NLP for process identification, discovery, and analysis & 33 papers (2009--2016) & Categorizes which phases of BPM life-cycle are covered, which NLP tools and techniques are used by various approaches. \\ 
	\cite{Indahyanti2022AutoGeneratingBP} & Process extraction from heterogeneous sources & 24 papers (2017-2022) & Primarily categorizes the kinds of input documents, ranging from event logs, to business rules, to textual descriptions, used for process extraction. \\
	\midrule 
	Our work & Text-to-model transformation approaches & \papers papers (2010-2023) & Compares approaches in a qualitative and quantitative manner. \\
	\bottomrule
	\end{tabularx}
\end{table}

The overview by \cite{Riefer16} aims to compare existing text-to-model transformation approaches, covering five of them in total, published up to 2015. In contrast to our qualitative and quantitative comparison, their work sticks to a qualitative perspective, in which they compare the input, output, and general method used by existing approaches.
As part of their work, the authors recognize the need for a systematic quantitative comparison using a set of textual descriptions and an accompanying gold standard, which is exactly what we provide in this work.

\cite{Maqbool18} particularly focuses on the extraction of BPMN models from textual descriptions, thus excluding other notations. Their analysis covers 36 papers on this topic, though it is important to recognize that these  also, inexplicably, include secondary articles, such as other surveys (e.g.,\cite{Riefer16}). Their work analyzes which NLP techniques (e.g., stemming, part-of-speech tagging) and tools are used by different approaches, and which BPMN elements the works support. Unlike our work, \cite{Maqbool18} does not compare approaches, but focuses on categorization.

The work by \cite{Bordignon2018NaturalLP} takes a broader perspective than just text-to-model transformation, particularly focusing on approaches that use NLP for the three main phases of the BPM lifecycle \citep{Dumas0031128}, i.e., process identification, discovery, and analysis. 
Their review categorizes which phases of the BPM life-cycle are covered, and which NLP techniques and tools are used  by a total of 33 papers, published between 2009 and 2016. Again, no comparison, qualitative or quantitative, is performed.

Finally, \cite{Indahyanti2022AutoGeneratingBP} analyze a broad range of approaches to extract processes from different kinds of input documents, ranging from structured business rules, to event logs, to textual process descriptions. Their work primarily focuses on a categorization of the input used by their selected approaches, thus not providing a comparison.

As shown in \autoref{tab:rwoverview}, our work differs considerably from these existing overview papers, primarily due to the depth of the analysis we perform. The majority of the overview articles focus on categorization of approaches, aiming to establish insights across the population of works they cover, such as that 11 approaches use the Stanford parser~\citep{Maqbool18}. By contrast, we focus on a detailed investigation and comparison of individual approaches. In that regard, as well as in its sole focus on text-to-model transformation, our work is closest to \cite{Riefer16}. However, in comparison to that work, we consider more and newer works (they cover five papers, up to 2015), analyze the evaluation of the approaches in depth by means of a qualitative evaluation, and provide a quantitative comparison of the tools on a common pipeline. To the best of our knowledge this latter task has never been systematically performed in the process extraction from text research area.

\section{Paper Selection Methodology}
\label{sec:protocol}
%\

The goal of this paper is to tackle the fragmentation of the process extraction research field by proposing a qualitative and a quantitative comparative analyses of well-known recent works, to better understand their contributions and limitations and to identify key challenges for the research community to address. The first problem to address in doing that is the choice of the procedure to select the works (hereafter primary studies) to be included in the comparative analyses. 

\mypar{Initial collection}
As a starting point, we selected all papers covered by four surveys discussed in \autoref{sec:related-works}, which yielded a set of 192 unique papers. 
Since these works primarily cover works published until 2018\footnote{\cite{Indahyanti2022AutoGeneratingBP} covers publications until 2022, but only part of it relates to process model extraction from text.}, we decided to augment our initial collection with papers that were since then published at the two flagship conferences for research on business process analysis, i.e., the  International Conference on Advanced Information Systems Engineering (CAiSE) and the International Conference on Business Process Management Conference (BPM). We considered their proceedings from 2018 up to 2022, taking into account their (i) main track, (ii) workshops, (iii) forum, and (iv) demo/doctoral consortium/awards/industrial tracks.
This resulted in 910 additional candidate papers, yielding a total of 1102 unique publications to examine.

%Since our goal was not to provide an extensive systematic literature review (SRL), we did abstain to do that. Instead we decided to start from and exploit the SRLs and surveys on text-to-model transformation mentioned in the previous section and consider as candidate papers, the works  mentioned there. Since those surveys cover works until 2018, we also decided to analyse proceedings of the  International Conference on Advanced Information Systems Engineering (CAiSE) and of the proceedings of the Business Process Management Conference (BPM) from 2018 up to 2022. These two conferences were included because of their extensive coverage of process mining and process extraction topics and we did analyse papers from the proceedings of the (i) main track, (ii) workshops, (iii) forum, and (iv) demo/doctoral consortium/awards/industrial tracks.

%After removing duplicate entries, we got a set of 192 unique papers from the SRLs and surveys and 910 papers form the CAiSE and BPM proceedings, for a total of 1102 unique papers to examine. 

\mypar{Paper selection}
To move from this initial selection of candidate papers to the final set of primary studies, we 
followed the typical steps performed in systematic literature reviews (see e.g., \cite{Kitchenham07guidelinesfor}) and set up explicit inclusion and exclusion criteria.

\begin{table}[!htbp]
	\caption{Inclusion and exclusion criteria.}
	\label{tab:inclusionexclusioncriteria}
	\centering
		\resizebox{.8\textwidth}{!}{
		\begin{tabularx}{\textwidth}{p{1cm}X}
		\toprule
IC 1: & The paper proposes an extraction approach of process model information from textual process descriptions.\\
IC 2: & The paper targets the reference imperative and declarative process modeling languages BPMN, and \declare / DCR-graphs, respectively.\\
%IC 3: & The paper proposes an approach that is general, thus not tied to a specific domain. \\

IC 3: & The paper's main focus is on process extraction from text, not on an associated task, such as text-to-model comparison. \\
IC 4: & The paper contains an evaluation of the presented approach or at least an application of the approach on a use case.\\
\midrule
EC 1: & The paper has been published before 2010.\\
EC 2: & The paper is not available online.\\
EC 3: & The paper is not in English.\\
EC 4: & The paper either was not under peer review or it is a technical report.\\
EC 5: & The paper is almost the ``same copy'' of others of the same author(s).\\
EC 6: & The paper is not long enough to present a complete approach. \\
	\bottomrule
	\end{tabularx}
	}
\end{table}

We established inclusion  and exclusion  criteria, shown in \autoref{tab:inclusionexclusioncriteria}, to define the relevant criteria to assess the appropriateness of selected papers and select a set of relevant primary studies to review. 
In order to be included, a paper must satisfy all inclusion criteria (IC 1-- IC 4) and none of the exclusion criteria  (EC 1 -- EC 6).
The inclusion criteria primarily focus on the topical fit of papers, i.e., they should present a process model extraction approach from textual descriptions (IC 1) that yield a process model in an established notation (IC 2). The extraction approach should be the main focus of the paper (IC3) and should be evaluated or showcased in some way (IC 4).
The exclusion criteria primarily focus on excluding papers that do not meet up to quality standards and be accessible.
%The criteria focus mainly on removing not scientifically valid papers and papers that do not satisfy this survey's primary criterion, that is, the paper had to be focused on an extraction approach for business processes within a Business Process Management setting. 

We applied the inclusion and exclusion criteria to the 1102 papers that constituted our starting data collection, by manually inspecting the papers. First, we evaluated all the candidate papers against the criteria based on their title, keywords, and abstract, allowing us to omit papers that were clearly out of scope. Next, we evaluated the remaining papers according to their full content.
This yielded a total of 10 papers\footnote{The interested reader can find all the details of the 1102 initial papers and their marking w.r.t. the inclusion and exclusion criteria at
	\gitrepo.}, as listed in \autoref{tab:primary-studies}. These 10 are considered in detail in the next section.

%
%
%The 10 remaining papers were then evaluated on the basis of three quality criteria:
%\begin{enumerate*}[label=\textbf{QA \arabic*}:]
%	\item Is a well-defined methodology used in the study?
%	\item Is the study clearly positioned within the state-of-the-art landscape?
%	\item Is the goal of the study elucidated?
%\end{enumerate*}
%For each paper, we marked the satisfaction of quality criteria QA1 -- QA3 with three possible scores: 1 (\textit{Yes}), 0 (\textit{No}), and 0.5 (\textit{Partially}). Finally, we selected only papers that reached at least a score of 2 out of the maximum possible score of 3.
%All 10 papers did satisfy the quality assessment, and they therefore define the set of primary studies used to perform our analysis.

%The interested reader can find all the details of the 1102 initial papers and their marking w.r.t. the IC/EC as well as the quality assessment at  
%\url{github.com/patriziobellan86/PETSotaAndChallenges}.
\begin{table}[!htbp] 
	\begin{center}
		\caption{The selected primary studies.}
		\label{tab:primary-studies}
	\resizebox{.8\textwidth}{!}{\begin{tabularx}{\linewidth}{p{5cm}Xp{3.5cm}} 
				\toprule
				\textbf{Reference} &  \textbf{Title} & \textbf{Venue }\\
				\midrule 

\cite{GoncalvesSB11} &  Let Me Tell You a Story -- On How to Build Process Models  & Journal of Universal Computer Science \\

\cite{Friedrich11} & Process Model Generation from Natural Language Text & CAiSE Conference \\

\cite{EpureMHDS15} & Automatic process model discovery from textual methodologies & RCIS Conference \\

\cite{Ferreira17} &A Semi-automatic Approach to Identify Business Process Elements in
Natural Language Texts &  ICEIS Conference\\

\cite{HonkiszK018} & A Concept for Generating Business Process Models from Natural Language
Description & KSEM Conference \\ 

 \cite{derAaDiCiccio19} & Extracting Declarative Process Models from Natural Language & CAiSE Conference \\ 
 
 \cite{DBLP:conf/bpm/QuishpiCP20} & Extracting Annotations from Textual Descriptions of Processes & BPM Conference \\ 
 
\cite{DBLP:conf/caise/0003W0LLZLW20} & An Approach for Process Model Extraction by Multi-grained Text Classification & CAiSE Conference \\ 

\cite{DBLP:conf/caise/AckermannNJ21} & Data-Driven Annotation of Textual Process Descriptions Based on Formal
Meaning Representations & CAiSE Conference \\ 

\cite{DBLP:conf/caise/LopezSNM21} & Declarative Process Discovery: Linking Process and Textual Views & CAiSE Forum \\
\bottomrule 
			\end{tabularx} %
		}
	\end{center}
\end{table}

%
%\begin{table}[tb] 
%\begin{center}
%	\caption{The primary studies selected.}
%	\label{tab:primary-studies}
%	\renewcommand{\arraystretch}{1.2}
%	\resizebox{\textwidth}{!}{\begin{tabular}{lll} 
%		\toprule
%		\textbf{Year} & \textbf{Conference Reference} & \textbf{Journal Reference}  \\ 
%		\toprule
%		2011 & &\citet{GoncalvesSB11}  \\ 
%		2011 & \citet{Friedrich11} & \\ 
%		2015 & \citet{EpureMHDS15} & \\
%		2017 & \citet{Ferreira17} & \\
%		2018 & \citet{HonkiszK018} & \\ 
%		2019 & \citet{derAaDiCiccio19} & \\ 
%		2020 & \citet{DBLP:conf/bpm/QuishpiCP20} & \\
%		2020 & \citet{DBLP:conf/caise/0003W0LLZLW20} & \\
%		2021 & \citet{DBLP:conf/caise/AckermannNJ21} & \\
%		2021 & \citet{DBLP:conf/caise/LopezSNM21}  & \\ 
%		\bottomrule
%	\end{tabular}
%	}
%\end{center}
%\end{table}

\section{Summary of the Papers}
\label{sec:process-extraction-from-text}

In this section, we briefly summarize the content of the \papers primary studies considered in our analysis.  Before describing the papers in detail, we provide an overview of the two high-level approaches that the proposed text-to-model solutions follow, which we use as a main categorization criterion for their presentation in the remainder.

\subsection{High-level Approaches for \PET}
\label{sec:approaches}

%%%%%%%%%%%%%%
\begin{figure}[tb]  % {img:text-to-model-abstract-view}
    \centering
    \includegraphics[width=0.6\textwidth]{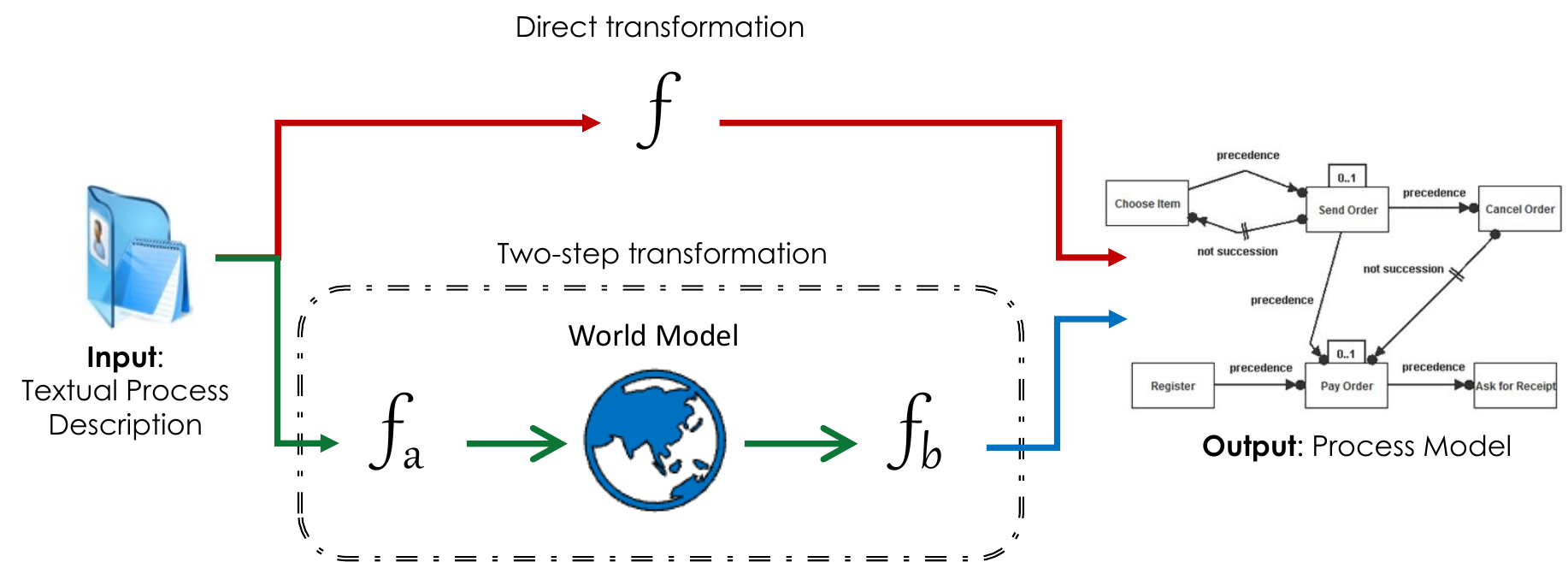}
    \caption{Two approaches to perform \pet.}
    \label{img:texttomodelabstractview}
\end{figure}

The 10 proposed solutions for \pet can broadly be divided into two main categories: direct and two-step transformation.  Solutions in the first category aim to \textit{directly map} a process description into its process model representation via a single function $f$, as graphically depicted in the top part of \autoref{img:texttomodelabstractview}\footnote{We have chosen Declare as an illustrative example of process model but the approach is agnostic to the specific modeling language.}. As shown in the remainder of the paper, function $f$ is typically implemented via a \textit{complex and ad-hoc tailored pipeline}. This approach has the advantage of defining a tailored transformation that can take into account all available contextual information that can help solve the problem. Nonetheless, this advantage becomes a drawback when we need to devise general solutions or when the algorithmic function $f$ is applied to different contexts. A further approach toward the implementation of a direct mapping $f$ is the exploitation of artificial neural networks. However, the huge quantity of data required to learn a model, and the small quantity of data available in this research domain make this strategy rarely adopted.

The second approach found in literature performs a \textit{two-step transformation} approach  to extract and create a process model, using a \textit{world model} as an intermediary representation.
As illustrated in the bottom part of \autoref{img:texttomodelabstractview}, the algorithmic function $f$ is here considered as a \textit{compound} function $\fa \circ \fb$: first, function $\fa$ extracts process elements from text and populates the intermediate representation, then function $\fb$ builds the process model diagram starting from the structured representation of the elements contained in the world model.

Note that $\fa$ and $\fb$ can be further broken down into smaller tasks that allow to better handle the problem complexity. For example, $\fa$ could contain a module that takes care of the resolution of anaphoric references within the text, another that filters out uninformative textual fragments that may act as noise in the later stages, and then specific modules tailored to the extraction of instances of specific process elements (e.g., activities or roles). Similarly, $\fb$ may contain modules devoted to the addition of process elements not explicitly described in the text (but conceptually required to create a correct model),
modules devoted to the connection of process elements together following the same logic conveyed in the textual description, or modules devoted to the generation of the labels of the different elements in the diagram.

\subsection{Primary Study Summaries}
\label{sec:brief-summary-of-the-papers}

\begin{table}[tb] % {tab:summary}
\begin{center}
	\caption{Summary of the approaches and techniques used.}
	\label{tab:summary-papers}
	\renewcommand{\arraystretch}{1.2}
	\resizebox{\textwidth}{!}{
		\begin{tabular}{@{}llllll@{}} 
	\toprule
		
		\textbf{Paper} & \textbf{Approach} & \textbf{Main Technique} & \textbf{Output} \\ 
		\toprule
		\citet{derAaDiCiccio19} & $f$ & Rules and Templates  & Declare \\ 		
		\citet{DBLP:conf/caise/LopezSNM21} & $f$ &  	Rules+Artificial Neural Network &  DCR graph  \\
		\arrayrulecolor{gray}
		\cmidrule{2-4}
%		%%%%%%%%%%%%%%%%%%%%%%%%%%%%%%%%
		
		\citet{GoncalvesSB11} & $\fa \circ \fb$  & 
		Regex Grammars, Rules and Templates  & BPMN \\ 
		\citet{Friedrich11}  & $\fa \circ \fb$ & Rules & BPMN \\ 
		\citet{EpureMHDS15} & $\fa \circ \fb$ &  Rules & BPMN \\ 
		\citet{HonkiszK018} & $\fa \circ \fb$ &  Rules & BPMN \\ 
		\arrayrulecolor{gray}
		\cmidrule{2-4}
%		%%%%%%%%%%%%%%%%%%%%%%%%%%%%%%%%
		\citet{Ferreira17} & {$f_a$} & Rules & 		Tagged entities \\
		\citet{DBLP:conf/caise/0003W0LLZLW20} & $f_a$ & Artificial Neural Network &Tagged entities \\
		\citet{DBLP:conf/bpm/QuishpiCP20} & $f_a$ & Rules & Tagged entities \\
		\citet{DBLP:conf/caise/AckermannNJ21} & $f_a$  & Artificial Neural Network & Tagged entities  \\ 

		\arrayrulecolor{black}
		\bottomrule
	\end{tabular}
	}
\end{center}
\end{table}

This section summarizes the \papers primary studies considered in our comparison, categorized according to the two high-level approaches described above. 
For each primary study, we briefly describe: (i) the aim of the paper; (ii) the proposed transformation approach; and (iii) a summary of the evaluation performed and the dataset used (if any). An overall summary of the papers is provided in \autoref{tab:summary-papers}, where  {$f$} means direct transformation, $\fa \circ \fb$  two-step transformation, and $f_a$ means that only the first part of the two-step transformation was performed. %, finally, the output specifies the language in which a process model is extracted or the fact that the procedure extracts only entities from text without composing them into a model.
  
\subsubsection{Direct Transformation Approaches} % (fold)

As shown in \autoref{tab:summary-papers}, two primary studies propose direct transformation approaches, both targeting declarative process models. 

\mypar{\citet{derAaDiCiccio19}}
The approach by \citet{derAaDiCiccio19} focuses on the extraction of declarative constraints from individual input sentences.
The discovered constraints are represented in the Declare language and are extracted through a tailored NLP pipeline. The work targets several challenges related to the discovery of \elements, such as the recognition of noun-based activities and the detection of different \emph{constraint restrictiveness} due to modal verbs.

The approach consists of three main steps: linguistic processing, activity extraction, and constraint generation. After a typical text normalization step of the input, \textit{linguistic processing} begins by extracting semantic components from the typed dependency relations representation of the sentence, aiming to identify key verbs, subject and objects, as well inter-relations between these terms. \textit{Activity extraction} then uses these extracted components to detect both verb-based (e.g., ``\textit{create order}'') and noun-based activities (e.g., ``\textit{order creation}''). 
Finally, \textit{constraint generation} establishes one or more Declare constraints based on the identified activities and their semantic inter-relations, allowing for the automatic extraction of five constrained types, as well as their negated forms.

The approach is evaluated on a collection of 103 sentences (i.e., constraint descriptions) and a corresponding gold standard, achieving an overall precision of 0.77, recall of 0.72, and F1 score of 0.74.
%The method achieved an overall precision of 0.77 and a recall of 0.72, yielding an overall F1 score of 0.74 on a text collection of 103 pairs consisting of a constraint description and its corresponding declarative constraints.

%%%\citet{DBLP:conf/caise/LopezSNM21}
%%%---------------------	
%(i) the settings of the research,
%(ii) the procedure presented, 
%(iii) the evaluation performed to test the ability of the proposed framework, 
%(iv) the dataset adopted to perform the evaluation step if it is described

\mypar{\citet{DBLP:conf/caise/LopezSNM21}}
The paper by \citet{DBLP:conf/caise/LopezSNM21} aims at automatically extracting declarative models from text, where the declarative models are represented using Dynamic Condition Response (DCR) Graphs.  Furthermore, the approach aims at maintaining links between the contents of a textual description and the corresponding DCR graph.

The approach is based on a combination of deep learning solutions and rules, called \textit{expert systems}. More in detail, the discovery problem is decomposed into two separate parts: a sentence classification problem and a Named Entity Recognition (NER) problem.
The two problems are tackled by means of two different BERT language models, which perform  sentence-level and word-level analysis of the input text, respectively.
The two language models parse a text sequentially, extracting roles, events, and relations to build the DCR graph in an  incremental manner.
The pre-trained language models are fine-tuned during the training phase and then used as predictors to discover process entities and relations from text.  

Training and evaluation are performed in a 10-folds cross-validation manner to assess the internal validity.
% of the proposed approach.
The external validity is assessed during the experimental evaluation stage by testing a dataset of 15 process descriptions, not used during training.
The authors demonstrated that the best combination of artificial neural networks and expert systems is able to achieve an average F1 score of 0.71. % for the problem of discover process elements and 

\subsubsection{Two-step Transformation Approaches (Realizing $\mathbf{\fa \circ \fb}$)} 

Four of the considered papers implement the entire $\fa \circ \fb$ pipeline for two-step transformation:

%Eight papers relate to the two-step transformation approach: four of them implement  the entire $\fa \circ \fb$ pipeline whereas the other four only focus on the extraction of process elements from text, thus only implementing $\fa$. 

%We present here the 8 papers that follow the two steps approach. First, we present 4 papers that cover the entire \pet function and provide both $\fa$ and $\fb$, then we describe the 4 papers that focus only on the extraction (identification) of structured process elements from the text. These latter papers can be considered works that only realize the first part of the transformation (that is part of $\fa$). 

%\todo{Han: this paragraph thing looks odd to me. We can also just create two subsubsections, one for the full approaches, one for the Fa ones}

%\subsubsection{Approaches realizing $\fa$ and $\fb$} % (fold)

\mypar{\citet{GoncalvesSB11}}
The work by \citet{GoncalvesSB11} presents \textit{storytelling mining}, an approach that aims at generating a BPMN representation of a process, starting from process instances represented as stories composed by users. 
When several alternative process models are generated, they are all presented to the process modeler. An analyst then has to combine them and validate the final version manually. 

The central idea of the approach is to apply text mining, information extraction, information retrieval, and natural language processing techniques in association with contextual elements of the collaborative told stories, to deal with this specific form of input text. 
First, the tellers record the stories in a repository. Stories are classified and clustered together, to group those referring to the same process. 
Then, a filtering stage excludes uninformative sentences from further analysis. 
A syntactical analysis of the informative sentences is performed through two different \textit{regular expression grammars} G.I and G.II.
The two grammars aim to find out \textit{noun phrases} (NP) that could correspond to actors and/or artifacts, and \textit{verb phrases} (VP) that could describe activities. 
In addition, a list of \textit{trigger words} is checked to discover gateways.
The process elements extracted from the text are memorized into an intermediate representation that adopts the CREWS~\citep{Achour98} model.
Finally, the process model is built by exploiting \textit{templates} starting from the intermediate representation. 
%When several alternative process models are generated, they are all presented to the process modeler. The analyst has to combine them and validate the final version manually.

The experimental evaluation measures the ability of the two regular expression grammars G.I and G.II to retrieve activities from two real-world case studies. The coverage of G.I is increased by coupling it with a list of trigger words, whereas G.II is used without trigger words.
The evaluation reveals that G.I has an accuracy of 45\%, while G.II performs better, with an accuracy of 78.57\%.

%%%\citet{Friedrich11}
%%%---------------------	 
\mypar{\citet{Friedrich11}}
The \pet approach proposed by \citet{Friedrich11} aims at extracting a BPMN model from a textual description. 

The procedure begins with a \textit{sentence-level analysis} phase to perform typical NLP pre-processing tasks such as text cleaning.Then, the pre-processed text is parsed by Stanford CoreNLP~\citep{CoreNLP} to obtain a semantic tree representation of the sentences. This representation is %Trees are 
used to determine if the verb of the sentence is active or passive and to extract instances of \emph{Actor} and \emph{Action}. 
%During this stage, \textit{example sentences} are filtered out using a keywords-based list.
%
Then, a \textit{text-level analysis} phase takes place to analyze the entire process description. 
During this phase, the co-references and their relative references are resolved. Conditional markers are checked against a list of conditional indicators to determine gateways. Here, \textit{WordNet}~\citep{wordnet} and \textit{VerbNet}~\citep{verbnet} are used to deal with the writing style variability of process descriptions.
The process entities and relations extracted are memorized into an intermediate representation and from that the process model is generated by means of rules.

%To evaluate the approach, t
The evaluation is performed on a publicly available dataset of 47 textual process descriptions, paired with the  corresponding process models. 
%In the experimental evaluation, the process model diagrams generated from texts are compared to the gold-standard ones. 
The authors demonstrate that this approach is able to yield an average graph edit distance similarity of 77\% between the BPMN process models extracted from text and the corresponding gold standard ones.

%47 pairs composed of a \processdescription, and a \diagram for performing the evaluation.
%The diagrams extracted with this prototype scored an average graph edit distance similarity of 77\% compared to the proposed gold-standard ones.

%%%\citet{EpureMHDS15}
%%%---------------------
\mypar{\citet{EpureMHDS15}}
The work described in \cite{EpureMHDS15} propose \textit{Text Process Miner}, an approach to mine a process model archaeological reports. This work differs from the others because the input text are multiple descriptions of process executions (in the archaeological domain) rather than the description of a process model. Thus, this work first extracts structured textual descriptions of single process executions from an unstructured text and then it leverages these structured textual descriptions to build the general BPMN process model.

The approach proposes a combination of natural language processing techniques to structure the input text and a set of rules to perform the extraction of the process instances. A special effort is devoted to the handling of active and passive forms of verb clauses. The pipeline starts by removing uninformative sentences and normalizing the text.
Then, a syntactic analysis aims at discovering activities and their relations, to create the structured textual descriptions of single process executions. Finally, a set of rules is applied to the structured textual descriptions to build the general BPMN process model.

The experimental evaluation is performed on a single archaeological study report and achieves a precision of 0.88. Moreover, a qualitative analysis performed by experts demonstrates that the BPMN process model generated is representative of the actual process followed in the archaeological domain.

%%%\citet{HonkiszK018}
%%%---------------------
\mypar{\citet{HonkiszK018}}
The \pet approach proposed by \cite{HonkiszK018} present a two steps transformation approach to automatize the discovery of BPMN models from natural language textual documents.

The procedure starts with a syntactic analysis of the input to extract Subject-Verb-Object constructs (SVO). 
Then, a semantic analysis is performed to further enhance the extraction by filtering out unnecessary SVO constructs during the discovery step. 

The actors are extracted from the text after an analysis of the dependency relations in search for nominal subject and nominal subject passive, conjunction dependencies, and after a keyword-based search for pronouns, relative pronouns, and hypernyms that belong to a specified list of admissible hypernym likes \textit{person, organization}. The SVO triplets are searched for in the parsed tree. The results are stored in a spread-sheet-like world model which acts as a structured intermediate representation of the entities and their connections, discovered in the text.
In particular the SVO triplets are rows in the spreadsheet world model in which the subject (the activity) and the possible object are recorded. Gateways are extracted by exploiting a keywords-based extraction. Finally, the BPMN process model is built up from the intermediate representation.

The paper lacks a proper experimental evaluation.\footnote{The authors state that they have tested the proposed approach against a test set gathered from academic sources. However, the details of the evaluation and the results are not reported in the paper.}  
Instead, the paper illustrates a use case of a single BPMN process model obtained from a textual description with the application of the  proposed approach.

\subsubsection{Process Element Extraction Approaches (Realizing $\mathbf{\fa}$ only)} 

Finally, four of the papers focus on just the extraction of process model elements from text, thus implementing function $\fa$ from the two-step transformation approach:

\mypar{\citet{Ferreira17}}
The work by \citet{Ferreira17} presents an approach to extract activities, events, and gateways from  unstructured textual sources such as reports, manuals, and norms.

The work exploits a rule-based approach, grounded on mapping rules. The procedure begins with a pre-processing step to normalize the input, and a syntactic analysis of the text to determine the complex structure of information beyond its pure surface realization.
The authors exploit the verb tense to differentiate an activity (present or future tense) from an event (past or present perfect tense).
Then, the text is analyzed to extract syntactic features such as tokenization, sentence recognition, part of speech tagging, lemmatization, dependency parsing, and named entity recognition. 
A set of 33 rules is then applied to this feature based representation to map the text into process elements, expressed in BPMN.
To increase the coverage of the extraction step, lists of signal words are checked to determine the presence (and type) of gateways.

%RIFORMULARE
%The experimental evaluation validates the rules exploited with an empirical evaluation that involves experts in process modeling.
%The evaluation is performed using a qualitative and a quantitative evaluation. 
%The first concerns an 
%A quantitative evaluation assesses the likelihood of the rules employing process modeling experts and providing positive results. 
A quantitative evaluation assesses the ability of the rules to correctly extract a small set of process elements from a testing dataset that comprises  56 texts, built by the authors, using manuals and documentation. 
The evaluation demonstrates that the rules exploited are able to achieve an overall F1 score of 0.87 on the extraction task.

%--------------
\mypar{\citet{DBLP:conf/caise/0003W0LLZLW20}}
The recent work of \citet{DBLP:conf/caise/0003W0LLZLW20} aims at classifying sentences in procedural texts by tagging them with the Activity, Sequence flow, Actor, and Artifact tags.

The approach is based on a hierarchical neural network, called Multi-Grained Text Classifier (MGTC), and has the advantage of addressing the problem without engineering any features.

The artificial neural network architecture adopted in this work is composed of four main parts: 
\begin{enumerate*}[{(i)}]
	\item an embedding layer to input the word's semantics in form of semantic embeddings;
	\item a recurrent encoding layer (bi-LSTM) to input an entire sentence to the higher layers;
	\item a convolutional layer adopted to extract meaningful features from the input sentence; and finally, 
	\item three classification layers that perform the three classification tasks: \textit{sentence classification}, \textit{sentence semantics recognition}, and \textit{semantic role labeling}. 
\end{enumerate*}

An experimental evaluation is conducted on two publicly available datasets, built by the authors, starting from a collection of cooking recipes and maintenance manuals. 
The evaluation aims at testing different configurations of neural networks on the three classification tasks mentioned above.
The proposed approach outperformed the pattern-matching baseline, yielding an accuracy of around 90\% on the three tasks in the two datasets.
%The authors demonstrated that the proposed approach is able to perform better than the pattern-matching-based method used as a baseline.
%report a statistically significant high score of the system in the two datasets compared to a pattern-matching-based method used as a baseline.

\mypar{\citet{DBLP:conf/bpm/QuishpiCP20}}
In another contribution, \citet{DBLP:conf/bpm/QuishpiCP20} propose a novel approach to automatically annotate process model entities and their relations expressed  with Annotated Textual Descriptions of Processes (ATDP) tags, within textual descriptions of processes.
%The approach aims to% that follow a declarative approach. 
%
%The proposed tool 
%annotates \processdescription with Annotated Textual Descriptions of Processes (ATDP) tags. 
Since ATDP annotations can be translated into linear time temporal logic over finite traces, the transformation of text into logical constituents opens up the possibility of applying formal reasoning to it. 

The strategy adopted in this approach couples the analysis of natural language with T-regex rules.
T-regex is a regular expression language that allows for defining patterns on the dependency tree of a sentence.
This pattern language has the advantage of being robust to variations in writing style and word use.
The transformation procedure begins with common pre-processing and normalization of the input text.
Then, a morphological analysis determines the grammatical category of each word and a \textit{name entity recognition} (NER) component detects person, location, organization, time, and numerical expressions in the text.
A word disambiguation step reduces the problem of semantic ambiguity of words, using WordNet \citep{wordnet} as a reference resource.
The information gathered in the previous steps is fed into a dependency parser to obtain a semantic tree representation of the input. Next, a \textit{semantic role labeling} component extracts theta roles (such as the agent, patient, and recipient). After these steps of analysis, the syntactic and semantics of the text are well exposed and T-regex rules are applied to extract process entities and their relations.

The experimental evaluation focuses on the extraction of activities and their relations.
The activity extraction task is performed on a collection of 18 process descriptions and 
%, and \diagram manually annotated to obtain a gold standard dataset.
the results show that the proposed approach achieves an overall F1 score of 0.71.
% and that the tool performs better in this task than a baseline was taken from~\citep{Friedrich11}.
%
The relation extraction task is evaluated in comparison to \cite{derAaDiCiccio19}.
A second dataset, only partially available, was created for this evaluation.
The results show that the proposed approach achieves an overall F1 score of 0.61.%, which is greater than the baseline.

%%%---------------------
%% Ackermann
%%%---------------------	
%(i) the settings of the research,
%(ii) the procedure presented, 
%(iii) the evaluation performed to test the ability of the proposed framework, 
%(iv) the dataset adopted to perform the evaluation step if it is described

\mypar{\citet{DBLP:conf/caise/AckermannNJ21}}
In a recent paper, \citet{DBLP:conf/caise/AckermannNJ21} propose an approach, called \textit{UCCA4BPM}, to the problem of tagging process model elements in textual descriptions. 

The idea relies on the use of Semantic Parsing in combination with Graph Neural Networks.
The approach takes advantage of the combination of word embeddings and syntactic features (POS tags) to represent the input text. Moreover, the authors try to alleviate the ambiguity drawbacks of syntactic parsing by adopting the \textit{Unified Conceptual Cognitive Annotation} (UCCA) schema for the formal meaning representation of utterances.
%The adoption of the UCCA schema has several benefits: (i) the schema is applicable in a cross-linguistic setting; (ii) it is able to represent the semantics of a whole paragraph; and (iii) it is cognitively plausible since it is based on cognitive categories.
The procedure begins with a pre-processing phase for text normalization. Then, Semantic Parsing takes place. The text is transformed into a graph representation following the UCCA schema, and the graph representation of the text is enhanced with (i) the semantic representation of words using semantic embeddings, and (ii) syntactic information of words using POS tags.  Finally, the graph neural model predicts process model elements of the enhanced graph representation of the text.

The proposed approach is evaluated twice, once against the task proposed in \cite{DBLP:conf/bpm/QuishpiCP20}, using both the original dataset proposed in that paper and a novel dataset introduced by the authors, and once against the task proposed in \cite{DBLP:conf/caise/0003W0LLZLW20} using the original dataset proposed in that plus the novel dataset mentioned above. For each dataset, the task is evaluated in a  5-fold stratified cross-validation manner. %, and is evaluated w.r.t. three different evaluation scores: (i) \textit{exact match} where a match is a true match if and only if the entire span of words predicted matches the gold standard one, (ii) \textit{partial match} where a match is valid if at least one word predicted matches the gold standard annotation, and (iii) \textit{fragmented match} where each word is evaluated against the gold standard annotation.
The results show that UCCA4BPM largely outperformed the approach proposed in \cite{DBLP:conf/bpm/QuishpiCP20} in the novel dataset that the authors propose, but not on the dataset originally used by \citeauthor{DBLP:conf/bpm/QuishpiCP20}. They also show that UCCA4BPM outperformed the approach proposed in \citet{DBLP:conf/caise/0003W0LLZLW20} in both datasets.

 \section{Qualitative Comparison}
\label{sec:qualitative-comparison}

In this section, we provide a qualitative comparative analysis of the primary studies on three aspects:
\begin{enumerate*}[(i)]
	\item the main characteristics of each solution; 
%	, in theapproach and techniques adopted to tackle the problem;
	\item the type of \elements extracted from the input data; and	
	\item the experimental evaluation performed to evaluate the proposed framework.
\end{enumerate*}

\subsection{Main Characteristics}
\label{ssec:mapproach-and-techniques-used}

We first compare the primary studies by describing their approach, the main technique(s) used, the level of user involvement, the form of the input text, the adopted intermediate representation(if applicable), and the generated output, as summarized in \autoref{tab:main-strategy}. 
%Table~\ref{tab:main-strategy} summarizes these characteristics for the \papers papers. 

%.  move back to the original position    
    
\begin{table}[tb] % {tab:main-strategy}
\begin{center}
	\caption{Detailed view on the solutions and their characteristics.}
	\label{tab:main-strategy}
	\renewcommand{\arraystretch}{1.2}
	\resizebox{\textwidth}{!}{\begin{tabular}{@{}lllllllll@{}} %\begin{tabular}{l|ccc|ccccc}
		\toprule
		 &  &  &  & \multicolumn{3}{c}{\textbf{Input}} \\
		 \cmidrule(lr){5-7}
		\textbf{Paper} & \textbf{Approach} & \textbf{Main Technique} & \textbf{User inv.} &  \textbf{Text} & \textbf{Structure} & \textbf{Focus} & \textbf{Intermediate Repr.} & \textbf{Output} \\ 
		\toprule

		\citet{derAaDiCiccio19} & \multirow{2}{*}{$f$}& Rules and Templates  & No & 	Sentence & Unstructured & Process  & -- & \declare \\

		\citet{DBLP:conf/caise/LopezSNM21} & & 	Rules+Artificial Neural Network & Yes & Entire & Unstructured & Process  & -- & DCR graph  \\

		\arrayrulecolor{gray}
		\cmidrule{2-9}
		%%%%%%%%%%%%%%%%%%%%%%%%%%%%%%%%
		\citet{GoncalvesSB11} & \multirow{4}{*}{$\fa \circ \fb$}  & 
		%Regex Grammars, 
		Rules and Templates  & Yes & Entire & Unstructured & Process  & CREWS & BPMN \\ 
		\citet{Friedrich11}  & & Rules & No & Entire & Unstructured &Process    & CREWS & BPMN \\ 
		\citet{EpureMHDS15} & & Rules & No &  Entire & Semi-structured & Executions & Structured table & BPMN \\ 
		\citet{HonkiszK018} & & Rules  & No & Entire & Unstructured & Process   & Spreadsheet & BPMN \\ 
		\arrayrulecolor{gray}
		\cmidrule{2-9}
		%%%%%%%%%%%%%%%%%%%%%%%%%%%%%%%%
		\citet{Ferreira17} & \multirow{4}{*}{$f_a$} & Rules & No & Entire & Unstructured & Process   & -- &
		Tagged entities \\
		\citet{DBLP:conf/caise/0003W0LLZLW20} & & Artificial Neural Network & No & Entire & Unstructured & Process   & -- &
		Tagged entities \\
		\citet{DBLP:conf/bpm/QuishpiCP20} & & Rules & No & Entire & Unstructured &Process    & -- & Tagged entities \\
		\citet{DBLP:conf/caise/AckermannNJ21} &   & Artificial Neural Network & No & Entire & Unstructured & Process  & -- & Tagged entities  \\ 
		\arrayrulecolor{black}
		\bottomrule
	\end{tabular}
	}
\end{center}
\end{table}

The second column in the table presents the classification of the primary studies into the main \textbf{approaches} introduced in Section~\ref{sec:approaches}. As described in \autoref{sec:brief-summary-of-the-papers}, two of the primary studies focus on the direct transformation approach ($f$), four implement the entire two-step transformation pipeline ($\fa \circ \fb$), and four only target the extraction of process elements ($\fa$).

%The third column reports the type of \textbf{techniques} used in the different paper, which were described in an individual manner in Section~\ref{sec:brief-summary-of-the-papers}. 
The \textbf{techniques} used by the solutions can be divided in two main groups: rule-based and neural network techniques. 
Rule-based techniques are used in 7 out of \papers papers. Specifically, \citet{EpureMHDS15} and \citet{Ferreira17} apply rules to the semantic structure of the  sentences in the text; \citet{Friedrich11} and \citet{HonkiszK018} enhance rules  with a list of specific words; and \citet{DBLP:conf/bpm/QuishpiCP20} expresses rules in form of Tree-base patterns applied on top of the dependency tree of each sentence. \cite{derAaDiCiccio19} and \cite{GoncalvesSB11} pair a rule-based technique with a templates-filling one to represent the \elements extracted in their output \diagram. In particular, the work by \cite{GoncalvesSB11} uses templates to generate process model elements starting from the data memorized into the intermediate representation, while the work by \cite{derAaDiCiccio19} generates declarative relations between activities by filling templates with activities extracted from the input. 
Two papers purely use neural networks \citep{DBLP:conf/caise/AckermannNJ21, DBLP:conf/caise/0003W0LLZLW20}, using a fine-tuned language model to extract \elements for the specific task. Finally, \cite{DBLP:conf/caise/LopezSNM21} combine a neural network with a set of rules to reduce the number of false positives, while increasing recall of the overall solution.

Concerning the level of \textbf{user involvement}, eight out of \papers papers propose a fully automatic pipeline.
The other two,  instead, require a user to manually choose, correct, and validate the output data.
Specifically, \cite{GoncalvesSB11} provides a user with a collection of possible \diagrams, from which the user should choose and validate one.
%The output of the framework of \cite{GoncalvesSB11} is a collection of possible alternative \diagrams and the user is required to choose the correct one and validate it.
%Also, during the story creation (that is, the generation of the input data), the \textit{teller} (a person who describes a process) has to manually validate actors and characters of the story.
In \cite{DBLP:conf/caise/LopezSNM21}, the framework constantly tries to increase coverage and accuracy of the system by taking advantage of user feedback on the output. The role of the user is to select the correct expert system (set of rules to apply) to refine the prediction of the neural model.

%The next three columns in \autoref{tab:main-strategy} summarize the type of input, output, and intermediate representations (when applicable). 

For the \textbf{input}, \autoref{tab:main-strategy} shows that eight of 10 solutions accept an entire, unstructured text that describes an entire process at once. Exceptions to this are the approach by \cite{derAaDiCiccio19}, which analyzes individual sentences at a time (corresponding to individual declarative constraints), whereas \cite{EpureMHDS15} focuses on documents that describe process executions in a semi-structured manner.

%describes whether a proposed solution accepts  an entire portion of text or single sentences; whether the text is completely unstructured or some sort of structure is required; and whether the text describes the process model or single process executions. The majority of the solutions accept an entire \processdescription written in a completely unstructured way. The only exceptions are \cite{derAaDiCiccio19}, which analyses an unstructured process description one sentence at a time, and~\cite{EpureMHDS15}, which focuses on documents that describes process executions, as already reported in the paper summary. 

Four works realize the entire $f_{a}\circ f_{b}$ pipeline by adopting \textbf{intermediate representations} (a.k.a. world models).
% that can be grouped into two different sets: the first one is provided by the
 \citet{GoncalvesSB11} and \cite{Friedrich11} both employ the  CREWS~\citep{Achour98} world model, whereas \cite{EpureMHDS15} and \cite{HonkiszK018} both use a tabular representation. 
 
% , adopted by~\cite{GoncalvesSB11, Friedrich11}, the second one is provided by a less structured representation either in forms of structured table or in forms of spreadsheet, adopted in~\cite{EpureMHDS15} and \cite{HonkiszK018}, respectively.
    
  Finally, regarding the \textbf{output}, we distinguish between the contributions that propose a complete pipeline and those that only realize $f_a$. 
If we consider the complete pipeline, Table~\ref{tab:main-strategy} shows that the four two-steps approach papers aim to extract an \emph{imperative} process model expressed in the BPMN modeling language, while the two direct approach papers aim to extract a \emph{declarative} process model expressed either in \declare or in DGR Graphs. 
If we move to the papers that realize $f_a$ only, they produce an output that is typically a list of sentences tagged with process entities. Here, \cite{Ferreira17,DBLP:conf/caise/0003W0LLZLW20} are inspired by \emph{imperative} paradigms (and in particular BPMN), and thus tag the text with elements of those modeling languages, while  \cite{DBLP:conf/bpm/QuishpiCP20,DBLP:conf/caise/AckermannNJ21} aim to tag the text with declarative process knowledge.

\subsection{Extracted Process Elements}
\label{ssec:process-model-elements-extracted}

To compare the scope of the different solutions, 
Table~\ref{tab:process-elements} shows their coverage in terms of the different process elements that they aim to extract.
We group the elements into two subsets: those referring to the control flow 
%(activity, event\footnote{The term ``event'' is widely used in BPM often to denote different entity types. The interested reader is referred to~\cite{DBLP:conf/caise/AdamoFG20} for a discussion. In this paper, we use the term event to denote BPMN events.}, gateway, sequence flow, message flow, and generic relations between activities, here extracted only when referring to relations in declarative languages) 
and those referring to elements that relate to activities, in particular actors, and artifacts~\citep{DBLP:conf/aiia/AdamoBFGGS17}.  
Since the elements extracted depend on the modeling language, we have grouped the papers according to that. Since DCR Graphs focuses on a subset of \declare patterns, for the sake of simplicity we consider it here a variant of \declare. In the table, we use ``No'' to indicate that a process element is present in the target modeling language but is not extracted by the pipeline at hand. We instead use ``-'' to indicate that the element is not present in the target modeling language, and thus its extraction does not concern that particular paper. Since approaches exist to extend \declare with data, we have considered the extraction of actors and artefacts also for approaches aiming at declarative entities. Also, 
since a distinction between imperative (BPMN) and declarative forms of output can be made for the papers that implement $f_a$ only, as described at the end of Section~\ref{ssec:mapproach-and-techniques-used}, we use ``-'', ``Yes'' and ``No'' accordingly. 

\begin{table}[tb] % {tab:process-elements}
	\renewcommand{\arraystretch}{1.2}
    \begin{center}
     \caption{Process elements extracted from the input text.}
     \label{tab:process-elements}
    \resizebox{\textwidth}{!}{\begin{tabular}{@{}lccccccccc@{}} %\begin{tabular}{l|ccc|ccccc}
    \toprule
    & & \multicolumn{6}{c}{\textbf{Control Flow}} &  \multicolumn{2}{c}{\textbf{Participant}} \\ 
	\cmidrule(lr){3-8}\cmidrule(lr){9-10}
    \textbf{Paper} &  \textbf{Output} & \textbf{Activities} & \textbf{Events} & \textbf{Gateways} & \textbf{Sequence flow} & \textbf{Message flow} & \textbf{\declare Relations} & \textbf{Actors} & \textbf{Artifacts} \\ 
   \midrule
   
	\citet{GoncalvesSB11} &\multirow{4}{*}{BPMN} & \yes & \yes & \yes & \yes & \yes & -- & \yes & \yes \\
	\citet{Friedrich11}  & & \yes & \yes & \yes & \yes & \yes & -- & \yes & \yes \\
	\citet{EpureMHDS15} & & \yes & \no & \yes & \yes & \no & -- & \yes & \no \\ 
	\citet{HonkiszK018} & & \yes & \yes & \yes & \yes & \yes & -- & \yes & \yes \\
	\arrayrulecolor{gray}
	\cmidrule{2-10}
	\citet{derAaDiCiccio19} & \multirow{2}{*}{\declare} &\yes & -- & -- & -- & -- & \yes & \yes & \no \\
	\citet{DBLP:conf/caise/LopezSNM21} & & \yes & -- & -- & -- & -- & \yes & \yes & \no \\
	\arrayrulecolor{gray}

	\cmidrule{2-10}
	\citet{Ferreira17} & \multirow{2}{*}{Tags Imperative} & \yes & \yes & \yes &  \no & \no & -- & \no & \yes \\
	\citet{DBLP:conf/caise/0003W0LLZLW20} & & \yes & \no & \no &  \yes & \no & -- & \yes & \yes \\ 
	\citet{DBLP:conf/bpm/QuishpiCP20} & \multirow{2}{*}{Tags Declarative} & \yes & -- & -- & -- & -- & \yes & \yes & \yes \\
\citet{DBLP:conf/caise/AckermannNJ21} & & \yes & -- & -- & -- & -- & \no & \yes & \yes \\
    \bottomrule
    \end{tabular}
    }
\end{center}
\end{table}

Let us first focus on the elements describing the control flow of a business process. 
% 
%%%%% ACTIVITY %%%%%%%
% 
Since activities are the core element of a \diagram, they are, not surprisingly, always identified and extracted from a \processdescription in all the primary studies. 
It is interesting to note that all four works aiming at extracting BPMN diagrams retrieve the core structure of an imperative process model, composed of activities, gateways, and the sequence flow. All but~\cite{EpureMHDS15} also complement this core structure with events\footnote{The term ``event'' is widely used in BPM often to denote different entity types. The interested reader is referred to~\cite{DBLP:conf/caise/AdamoFG20} for a discussion. In this paper, we use the term event to denote BPMN events.} and message flow. 
All four declarative approaches, in turn, extract declarative relations.
%Control flow elements, in the form of \textbf{\declare relations}, are extracted by all four works that address declarative models or annotate text with declarative constructs. 

%%%%% ACTORS/ ROLES %%%%%
Finally, actors\footnote{We have classified as actors all mentions of actors and roles.} are also a rather popular element, while artifacts\footnote{We have classified as artifacts all mentions of artifacts, business entities, and (nonagentive) resources.} are slightly less present in the output of approaches that target the BPMN languages and completely absent in approaches that target the \declare language, possibly due to the still preliminary status of work that extend \declare with data. Both entities are rather popular in the four papers that only implement a partial pipeline, where they are second only to activities. 
% Regarding the strategy, the situation is similar to the extraction of the other \elements.
% In general, actors and artifacts are extracted using rules applied on top of the analysis of NPs and VPs of each sentence.
% \btext{Rules in form of patterns or in form of grammar expression rules are explored in \citep{GoncalvesSB11,Ferreira17}. In the contribution of \citep{HonkiszK018}, rules are extended by the use of word lists and a gazetteer. This step also makes use of WordNet.
% A more elaborated strategy is proposed in \citep{Friedrich11}.
% In this contribution, actors extracted are validated by checking the presence of terms against a word-list, and the analysis of the dependency relations between words.
% T-regex patterns are defined in the work in \citet{DBLP:conf/bpm/QuishpiCP20}.
% The extraction performed via artificial neural networks is explored in \citep{DBLP:conf/caise/AckermannNJ21}.
% }

\subsection{Conducted Evaluations}
\label{ssec:experimental-evaluation}

In this section, we compare the experimental evaluations conducted in the primary studies.
Since the work of~\cite{HonkiszK018} does not report an evaluation but only a sample application on a single use case text, we omit it from the comparison performed in this section.
%\todo{aggiungere esempio di altro task in footnote}
%
We analyze the papers by comparing: 
\begin{enumerate*}[(i)]
	\item what is evaluated;
	\item the dataset(s) used in the evaluation; 
	\item the metrics adopted; and 
	\item the result of the evaluation in terms of performance.
\end{enumerate*} 
The results are summarized in Table \ref{tab:experimental-evaluation}\footnote{Note that the papers which report F1 scores also report precision and recall values used to compute the F1 score. For the sake of presentation we have decided to report here only the F1. The interested reader can find the values of precision and recall in the different papers.}

%%%%%%%%%%%%%%%%%%%%%%%%%%%%%%%%%%%%%%%
\begin{table} % {tab:Testing}
\resizebox{.58\textwidth}{!}{
\begin{minipage}{\textwidth}
\begin{threeparttable}[tb]
	\caption{Overview of the evaluations conducted in the papers.} 
	\label{tab:experimental-evaluation}
\begin{tabular}{lm{5cm}m{5cm}clm{4cm}}
    \toprule
	 & & \multicolumn{2}{c}{\textbf{Test dataset}} \\
    \cmidrule(lr){3-4}
	 \textbf{Paper} & \textbf{What is tested} & \multicolumn{1}{c}{\textbf{Dataset}} & \textbf{Availability} & \textbf{Metric} & \textbf{Score}  \\ 
    \midrule
\rowcolor{lightgray!30}
\citet{GoncalvesSB11} &
	Regular Expression Grammars \textit{G1}, \textit{G2} for process model elements extraction & 2 cases studies (c.s.1 and c.s.2) &
	\no & Accuracy & \textit{G1}: c.s.1 33\%, c.s.2 45\%
	   \textit{G2}: c.s.1 80\%, c.s.2 78\%
	\\ % \midrule

\citet{Friedrich11} & Diagram similarity & 47 $\langle$process description, model diagram$\rangle$ pairs	& \yes & Graph Edit Distance	& 0.77 	\\

\rowcolor{lightgray!30}
\citet{EpureMHDS15} & Activity extraction & Archaeological report composed of 34 executions & \no & Precision & 0.88 \\

\citet{Ferreira17} %* checked
	&	Extraction of: (i) Activity; (ii) Event; (iii) Exclusive Gateway; (iv) Inclusive Gateway & 56 \processdescription & \no & F1 score & (i) 0.807; (ii) 0.868; (iii) 0.816; (iv) 0.927 \\

\rowcolor{lightgray!30}
\citet{derAaDiCiccio19} %* checked
	 & 	\declare pattern extraction  & 103 $\langle$ patterns description, patterns$\rangle$ pairs	& \yes & F1 score	& 0.74 	\\

\citet{DBLP:conf/bpm/QuishpiCP20} %* checked
&
Extraction of: (i) Activity; (ii) Relation  &  
	(i) subset of \citep{Friedrich11} + subset of \citep{DBLP:journals/dke/Sanchez-Ferreres18} (ii)subset of~\citep{Friedrich11} + subset of~\citep{DBLP:journals/dke/Sanchez-Ferreres18} +~\citep{derAaDiCiccio19}  & Partial & 
	 F1 score & (i) 0.71; (ii) 0.61 	
	\\
\rowcolor{lightgray!30}
\citet{DBLP:conf/caise/0003W0LLZLW20} %* checked
	 	 & (i) single sentence classification; (ii) sentence semantics; (iii) semantic role labeling
	 	 \tnote{1}
	 	 & A collection of recipes and manuals &
	\yes & Accuracy & (i) recipe 93.34\% \quad manual 91.74\%; (ii) recipe 91.53\% \quad manual 86.49\%; (iii) recipe 82.39\% \quad manual 80.44\% 	\\

\citet{DBLP:conf/caise/AckermannNJ21} & 
(a) extraction task of \cite{DBLP:conf/caise/0003W0LLZLW20}; (b) extraction task of \cite{DBLP:conf/bpm/QuishpiCP20}
 & (i) \cite{DBLP:conf/caise/0003W0LLZLW20}+(ii) \cite{DBLP:conf/bpm/QuishpiCP20}+ (iii) 5 texts from \cite{DBLP:conf/caise/AckermannNJ21} & \yes & F1 score  & 
	(a i) COR: 98.26\%; (a i) MAN: 97.78 \%; (a iii) 65.05;
	(b ii) 88.15 \%; (b iii) 90.66
  \\
\rowcolor{lightgray!30}
\citet{DBLP:conf/caise/LopezSNM21}  &  \declare pattern extraction  & subset of \cite{DBLP:conf/bpm/QuishpiCP20} subset of \cite{derAaDiCiccio19} + 5 texts of their dataset	& partial & F1 score	& 0.71~\tnote{2}		\\
\\

    \bottomrule
\end{tabular}
\begin{tablenotes}
	\item [1] Our understanding of the paper is that the \textit{single sentence classification} task classifies a sentence (or a textual fragment) as an Activity or not. 
 The \textit{sentence semantic} task classifies the semantics conveyed in a sentence (or in a textual fragment) into gateway, loop, end-event, and sequence flow.
 The \textit{semantic role labeling} task classifies the words of the text into three categories: role, action, and activity data.
  \item [2] We report the F1 score of the ``Best combination'' item obtained on the external validation test dataset.
\end{tablenotes}	
\end{threeparttable}
\end{minipage}
}
\end{table}
%%%%%%%%%%%%%%%%%%%%%%%%%%%%%%%%%%%%%%%%%%%%%%%%%%%%%%%%%%%%%%%%%%%%%%%%

The table shows a highly heterogeneous situation both in terms of what is evaluated, and the datasets and metrics used to perform the evaluation itself. This makes a direct comparison almost impossible. Nonetheless, we provide an attempt to comparatively analyze the different columns. 

%%%%% WHAT IS TESTED
Focusing on \textbf{what is tested}, we can divide the contributions into three main groups: 
\begin{enumerate*}[(i)]
	\item contributions that evaluate the quality of the diagram extracted from the input text; 
	\item contributions that evaluate the extraction of \elements, and 
	\item contributions that evaluate further aspects.
\end{enumerate*}

The works of \citet{Friedrich11}, \citet{derAaDiCiccio19} and \citet{DBLP:conf/caise/LopezSNM21} belong to the first group. They differ in the way the quality of the extracted diagram in is tested. Indeed, \citet{Friedrich11}  evaluates the diagram similarity between the extracted BPMN process model and the one provided as a gold standard, while the remaining two works evaluate whether the correct \declare pattern is extracted from the input sentence. 
The works by \cite{GoncalvesSB11,EpureMHDS15,Ferreira17,DBLP:conf/bpm/QuishpiCP20} and the work of \cite{DBLP:conf/caise/AckermannNJ21} for one of the extraction tasks belong to the second group. They evaluate the ability to extract different types of elements. It is easy to see that no common evaluation task emerges from these papers. Indeed with the exception of \citep{DBLP:conf/caise/AckermannNJ21} all papers introduce their own evaluation task (and datasets and metrics as we will see in the remaining of the section).  
\cite{DBLP:conf/caise/0003W0LLZLW20} and \cite{DBLP:conf/caise/AckermannNJ21}, for one of the extraction tasks, evaluate the ability to identify specific properties of the input sentence, e.g., the semantic role labeling.

%%%%%%%% TESTING dataset %%%%%%%%%%%%%%%%%%%     
Concerning the datasets (columns 3 and 4 of Table \ref{tab:experimental-evaluation}), it is easy to see that the different works were tested using different \textbf{datasets} or different sub-set of the same dataset. Together with the variety of evaluation tasks, this also makes a rigorous comparison extremely hard. 
Only four datasets are completely publicly available: the one proposed by \citet{Friedrich11}, for the extraction of BPMN process models, the one proposed by \citet{derAaDiCiccio19} for the extraction of \declare patterns, and the ones proposed by \citet{DBLP:conf/caise/0003W0LLZLW20} and \citet{DBLP:conf/caise/AckermannNJ21} which contains annotated sentences (each one with their own annotation schema).

%%%%%%%% METRIC %%%%%%%%%%%%%%%%%%%    
Focusing on the \textbf{metrics}, \citet{Friedrich11} adopts a graph-based metric to quantitatively evaluate the quality of the diagrams extracted from a textual description of a process in terms of diagram similarity.
All the other works adopt different and varied information retrieval metrics.

%%%%%%%% SCORES %%%%%%%%%%%%%%%%%%%   
As a consequence of the high heterogeneity illustrated so far, analyzing the \textbf{scores} in a unifying manner is a problematic task. Indeed even when the task appears to be similar it is difficult to compare the results due to e.g., the different evaluation metrics, or the datasets used. 
To make an example, the task of activity extraction is evaluated in 4 papers   \citet{GoncalvesSB11,EpureMHDS15,Ferreira17,DBLP:conf/bpm/QuishpiCP20}. Nonetheless, a direct comparison of the results is difficult since these works often adopt different metrics. Even in the case of \citet{Ferreira17} and \citet{DBLP:conf/bpm/QuishpiCP20}, which exploit the same metric, it is difficult to say if a higher score is due to a better tool or on the evaluation performed on input text with different levels of complexity. 
%s
Few works provide direct comparisons. In particular \citet{DBLP:conf/bpm/QuishpiCP20} provides a direct comparison with \citet{derAaDiCiccio19}, and \citet{DBLP:conf/caise/AckermannNJ21} directly compares their approach against the works of \citet{DBLP:conf/caise/0003W0LLZLW20} and \citet{DBLP:conf/bpm/QuishpiCP20}.

\section{Quantitative Comparison}
\label{sec:quantitative-comparison}

This section presents a quantitative comparison of the existing approaches for \pet, to complement the qualitative analysis from the previous section. Performing such a quantitative comparison is far from straightforward, though. The evaluation results reported in the primary studies  themselves cannot be directly compared, due to differences in the employed datasets and evaluation metrics, as highlighted in \autoref{tab:experimental-evaluation}. Furthermore, the output obtained by the different approaches cannot be compared either, due to the heterogeneity of the tasks that the works address (cf. \autoref{tab:process-elements}).

%Performing a quantitative comparison of the tools for \pet is not a straightforward task. Indeed the analysis summarised in Table~\ref{tab:experimental-evaluation} shows the unfeasibility of a direct comparison due to the heterogeneity of extraction tasks, evaluation datasets, and metrics used in the different evaluations.  

Therefore, in order to still be able to compare the approaches, we focus on a specific task that forms a crucial part for any \pet approach: the extraction of process model elements and their corresponding relations. 
%In order to compare the tools we decided therefore to focus on a specific task that can be considered as common to all the tools we were able to retrieve. This task is that of process model elements and relations extraction from text. 
In a way, this can be considered as performing only the $f_a$ step of the pipeline illustrated in Figure~\ref{img:texttomodelabstractview}. 
Here, we need to point out that not all approaches focus on this extraction task as their final output. Furthermore, it required us to adapt the code of the available approach implementations in order to isolate the methods that focus on the extraction of process model elements, which was particularly challenging for the tools that implement the entire transformation pipeline $f$. 
Nevertheless, we argue that these concerns are outweighed by the benefits, since in this manner we are able to perform a quantitative comparison of existing tools on a task that is a crucial component in any \pet pipeline and it also provides a basis for future comparison of newly developed approaches against \sota works.
%We are fully aware that not all the tools were designed to have this as their final output. We are also aware that manipulating the code so as to isolate the identification of process elements within the text was especially sensitive for the tools implementing the entire pipeline $f$ using a direct transformation approach but our choice was motivate by two main factors: first, this is a core task that all the tools that we identified  could execute, second it is a fundamental task in a \pet pipeline. Therefore, this is an appropriate and fair task that we can use here to compare the tools and also propose for the future tools to compare against \sota works. 

As a basis for this comparison, we first determined which of the \papers primary studies from the previous sections are suitable for comparison. 
The fundamental ingredients we needed to perform our comparative extraction task were: the tools, a way to adapt the tools to the specific extraction task we propose, an annotated dataset (that is, a gold standard) upon which to compare the tools, and the evaluation metrics to be used. 
If no implementation was listed in the paper, we contacted the authors and asked for the version described in the paper or an updated version if available. In May 2023 we finalized our list by excluding all the tools that were either not available, or for which we did not receive any working version or support to make them work. 
%Concerning the tools we first searched, for each of the \papers selected contribution, whether the tools linked in the papers were available or not. If they were not available we contacted the authors and asked for the version described in the paper or an updated version if available. In \btext{May 2023} we finalized our list by excluding all the tools that were either not available, or for which we did not receive any working version or support to make them work. 
In this manner, we obtained \tools executable tools, corresponding to the works by 
%As a result of this process, we were able to retrieve and make use of the \tools tools proposed 
by \citet{derAaDiCiccio19, GoncalvesSB11, Friedrich11, EpureMHDS15, HonkiszK018, DBLP:conf/caise/0003W0LLZLW20}. 

Next, \autoref{ssec:petdataset} presents the annotated \PETD~\citep{DBLP:conf/bpm/BellanADGP22} we used as basis for our quantitative comparison, \autoref{ssec:evaluation-pipeline} describes the evaluation task on which we compare the \tools approaches in detail, \autoref{ssec:quantitative-metrics} presents and motivates the employed evaluation metrics, and, finally, \autoref{ssec:results-relaxed-comparison} the results obtained in this manner. Raw results and the scripts used to obtain them can be found in our work's repository.\footnote{\url{https://github.com/patriziobellan86/PETSotaAndChallenges}}

%The way we did use the tools to perform the comparative extraction task is described in Section~\ref{ssec:evaluation-pipeline}. Concerning the annotated dataset we have selected a recent annotated dataset for \pet called \PETD~\citep{DBLP:conf/bpm/BellanADGP22}. To the best of our knowledge this is, in fact, the only publicly available dataset manually annotated with process model elements mentions and their relations and is briefly illustrated in section~\ref{ssec:petdataset}. Finally, the chosen standard information retrieval metrics of precision, recall, and F1 score are presented and motivated in Section~\ref{ssec:quantitative-metrics}).

\subsection{The \PETD}
\label{ssec:petdataset}

The \PETD \citep{DBLP:conf/bpm/BellanADGP22} (\petonly for short) is a recent dataset containing the manually-annotated versions of 45 textual process descriptions. The annotations provided in this gold-standard resource capture the core process model entities and their relations, as described in the texts. 
We here briefly introduce the subset of the \petonly annotation schema and its elements that are used our quantitative comparison, whereas we refer to the original paper \citep{DBLP:conf/bpm/BellanADGP22} and its associated repository for further details on the dataset and the employed annotation guidelines.

%The entire description of the dataset, the annotation guidelines, the annotation schema, and the annotation process are out of the scope of this paper. Here we briefly introduce the subset of the \PETD annotation schema and its elements that are used in this paper as they are the basis of the extraction tasks upon which we have decided to test the \tools tools. They cover the main components of \petonly, and the core elements of a business process, and are the basis of the extraction tasks upon which we have decided to test the \tools tools. 

 \begin{figure}[!htbp]
 \centering
   \includegraphics[width=0.5\textwidth]{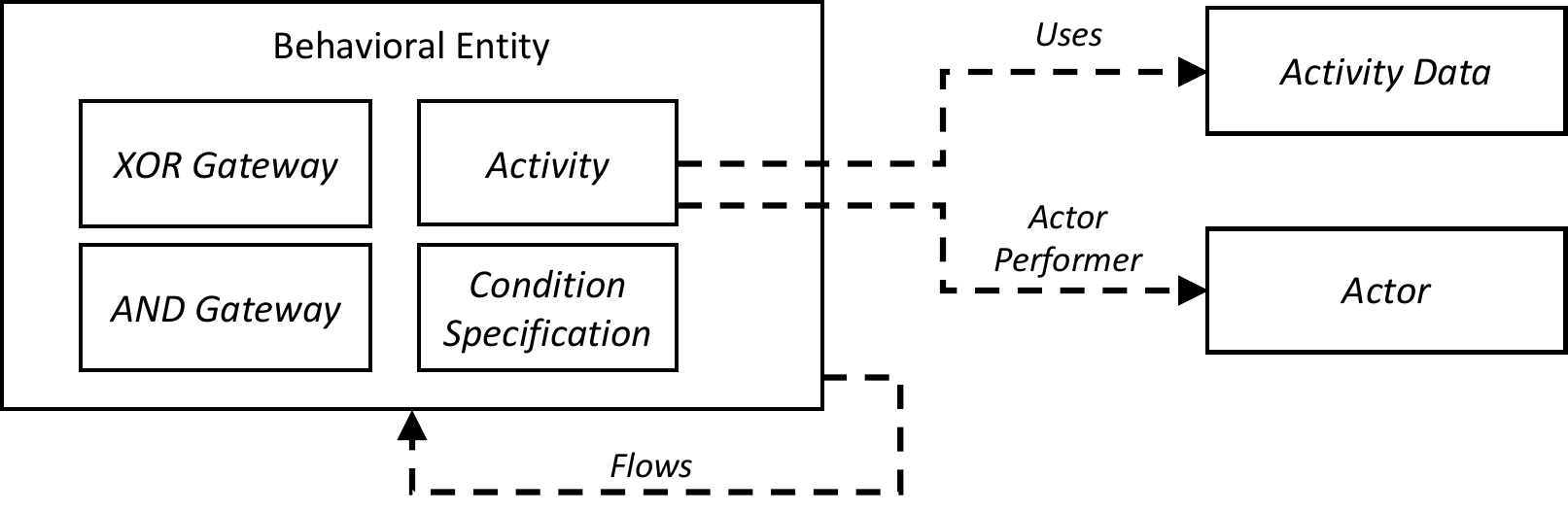}
 \caption{Extract of the PET dataset's annotation schema used in this research.}
 \label{fig:annotationschema}
\end{figure}

The central element of the \petonly annotation schema, illustrated in Figure~\ref{fig:annotationschema} is an \textit{Activity}, which represents a single task performed within a process (e.g., send an email). Differently from customary BPM terminology, in this schema, the activity is broken down into the activity entity, which captures only the expression of the ``action'' of a (classical BPM) activity in text, and the  \textit{Activity Data} entity, that captures the object of the action. These two entities are linked by the \textit{Uses} relation.\footnote{This choice allows for the annotation of activities where the action and object are separated from each other in the text, as well as for situations where multiple activities refer to the same data object.} Thus, in the \PETD, the sentence ``send an email'' would see ``send'' annotated with \textit{Activity}, ``email'' annotated as \textit{Activity Data}, and ``send'' and ``email'' would be linked by means of the \textit{Uses} relation.
An \textit{Actor}\footnote{In the original \PETD referred to as an \textit{Organizational} entity.} defines the organizational process participant involved in an activity execution (e.g., the customer office). The \textit{Actor Performer} relation connect an activity to the actor that performs/is responsible for the activity execution. 
Regarding the control flow structure, the schema captures mentions of parallel (\textit{AND Gateway}) and exclusive (\textit{XOR Gateway}) gateways in a text. %It also enables the capturing of the \emph{Same Gateway} relation when two or more different portions of text refer to the same gateway 
The \textit{Condition Specification} entity captures the condition a process model instance must satisfy in order to be allowed to enter a specific XOR branch of the control flow structure. Thus, in the \PETD, the sentence ``if age is greater than 35'' would see ``if'' annotated with \textit{XOR Gateway} and ``greater than 35'' with \textit{Condition Specification}. The control flow is defined as a sequential temporal succession of behavioral entities captured by means of the \textit{flow} relation. The reader can refer to \cite{DBLP:conf/bpm/BellanADGP22} for further details and examples. 

\subsection{Evaluation Task}
\label{ssec:evaluation-pipeline}

To define the extraction task in our quantitative comparison, we first mapped the process model elements that are extracted by the available approaches (cf. \autoref{tab:process-elements}) to the elements in the PET annotation scheme. 
%As the reader can easily see, the \PETD annotation schema illustrated in Figure~\ref{fig:annotationschema} contains entities and relations which do not correspond exactly to the list of process elements extracted by the different tools illustrated in the head of Table~\ref{tab:process-elements}. Therefore  the first step we had to do in order to define our evaluation task was to identify the \petonly entities and relations that correspond (in a single or composite form) to the elements extracted by the different tools and listed in Table~\ref{tab:process-elements}. 
This mapping is shown in Table~\ref{tab:correspondence}, which uses the symbols ``$\equiv$'' and ``$\subset$'' to respectively denote exact and partial correspondences, as explained in the following, and the symbol ``-'' to indicate an element not available in \petonly. 

 \begin{table}[!htbp]
	\begin{center}
		\caption{Correspondences between the elements extracted by the \papers papers (from Table~\ref{tab:process-elements}) and the \petonly annotation scheme.}
		\label{tab:correspondence}
		\resizebox{.6\textwidth}{!}{\begin{tabular}{lcl} 	
    \toprule
	\textbf{Table~\ref{tab:process-elements}} & \textbf{Correspondence} & \textbf{PET Element}\\
	\midrule
     Activity & $\equiv$ & $\langle$\emph{Activity} + \emph{Uses} + \emph{Activity Data}$\rangle$\\
	 Gateway  & $\supset$ & $\langle$ \emph{XOR Gateway} + \emph{Condition Specification}$\rangle$\\
	 Gateway & $\supset$ & \emph{AND Gateway}  \\
	 Sequence Flow & $\equiv$ & \emph{Flow} \\
	 Actor & $\supset$ & $\langle$\emph{Actor} + \emph{Actor Performer}$\rangle$\\
	 Event & & -- \\
	 Message Flow & & -- \\
	 DECLARE relation & & -- \\
	 Artifact & & -- \\
    \bottomrule
    \end{tabular}}
	\end{center}
    \end{table}

These correspondences are as follows:
\begin{itemize}[noitemsep]
	\item  The entity ``Activity'' as intended in Table~\ref{tab:process-elements} corresponds to the triple \emph{Activity}, \emph{Uses}, \emph{Activity Data} in \petonly;
	\item The generic ``Gateway'' entity mentioned in Table~\ref{tab:process-elements} is refined in \emph{AND} and \emph{XOR Gateway} in \petonly. For the \emph{XOR Gateway} \petonly distinguishes between the mention identifying a branch and and the condition that must be met for the execution to follow that branch.
	
	\item  The entity ``Sequence Flow'' in Table~\ref{tab:process-elements}, instead, corresponds exactly to the entity \emph{Flow} in \petonly
	
	\item The identification of ``Actor'' as in Table~\ref{tab:process-elements} typically enables us to identify both the Actor itself and the relation that the actor has with the activity it performs. Therefore, we map such Actor identification to both the \emph{Actor} and corresponding \emph{Actor Performer} relation in the PET schema.

%	 In fact, most tools extract an actor by identifying a portion of a sentence such as ``the customer sends the form''. This relation is in almost all the papers we considered of performing the activity. For this reason we did consider this relation in this paper. In general an Actor could be related to activities also with other forms of participation (e.g., it could be the one receiving the effect of an activity). For this reason we identify ``Actor'' in Table~\ref{tab:process-elements} only partially corresponding to the pair ``Actor, Actor Performer''. 
	
	\item 	Event, Message Flow, DECLARE relation, and Artifact are not present in \petonly and therefore not considered here. 
\end{itemize}

%As mentioned in Section~\ref{ssec:petdataset} the entity ``Activity'' as intended in Table~\ref{tab:process-elements} corresponds to the triple ``Activity, Uses, Activity Data'' in \petonly; the generic ``Gateway'' entity mentioned in Table~\ref{tab:process-elements} is refined in AND and XOR Gateway in \petonly. For the XOR Gateway \petonly distinguishes between the mention identifying a branch and and the condition that must be met for the execution to follow that branch. The entity ``Sequence Flow'' in Table~\ref{tab:process-elements}, instead, corresponds exactly to the entity ``Flow'' in \petonly; the identification of ``Actor'' as in Table~\ref{tab:process-elements} typically enables us to identify both the Actor itself and the relation the actor has with the activity. In fact, most tools extract an actor identifying a portion of sentence such as ``the customer sends the form''. This relation is in almost all the papers we did consider of performing the activity. For this reason we did consider this relation in this paper. In general an Actor could be related to activities also with other forms of participation (e.g., it could be the one receiving the effect of an activity). For this reason we identify ``Actor'' in Table~\ref{tab:process-elements} only partially corresponding to the pair ``Actor, Actor Performer''. 
%Event, Message Flow, DECLARE relation, and Artifact are not present in \petonly and therefore not considered here. 
Based on this mapping, Table~\ref{tab:pet-elements-relations-tools} shows the 
entities and relations from the PET annotation scheme considered by the \tools approaches.
%The reformulation of Table~\ref{tab:process-elements} in terms of the \petonly elements we were able to extract using the different tools is reported in Table~\ref{tab:pet-elements-relations-tools}. 
The elements that we were able to extract are marked with ``$\surd$'', those that were not meant to be extracted in the text tagging phase or not meant to be extracted at all are marked with ``-'', and, finally, those that were meant to be extracted but we were not able to obtain by using the tool are marked with ``NA''.   
One special consideration concerns the paper by \citet{DBLP:conf/caise/0003W0LLZLW20}, for which we were able to use the tool to extract the single entities Activity, Activity Data and Actor from the text, but it was not possible to use it for extracting the relations among them. 

 \begin{table}[tb]
	\begin{center}
		\caption{PET Process elements and relations extracted from each tool tested.}
		\label{tab:pet-elements-relations-tools}
	\end{center}
	\resizebox{\textwidth}{!}{\begin{tabular}{lccccccccc} 	
    \toprule
& \multicolumn{6}{c}{\textbf{PET Entity}} & 
  \multicolumn{3}{c}{\textbf{PET Relations}} \\ 
	\cmidrule(lr){2-7}\cmidrule(lr){8-10}
    \multirow{2}{*}{\textbf{Paper}} & 
\multirow{2}{*}{\textbf{\emph{Activity}}} & 
\textbf{\emph{Activity}} & 
\multirow{2}{*}{\textbf{\emph{Actor}}} & 
\textbf{\emph{XOR}} & 
\textbf{\emph{Condition}} & 
\textbf{\emph{AND}} &
\multirow{2}{*}{\textbf{\emph{Uses}}} & 
\textbf{\emph{Actor}} &  
\multirow{2}{*}{\textbf{\emph{Flow}}}  \\
%%%%%%
    &  &  
\textbf{\emph{Data}} &  &  
\textbf{\emph{Gateway}} & 
\textbf{\emph{Specification}} & 
\textbf{\emph{Gateway}} &  &    
\textbf{\emph{Performer}} & \\
\midrule 
\citet{GoncalvesSB11} &  
 	$\surd$ & $\surd$ & $\surd$ & NA  & NA  & NA  &
 	$\surd$ & $\surd$ & $\surd$\\
\midrule  
\cite{Friedrich11} & 
 	$\surd$ & $\surd$ & $\surd$ & $\surd$ & -  & $\surd$ &
 	$\surd$ & $\surd$ & $\surd$  \\
\midrule
\cite{EpureMHDS15} & 
 	$\surd$ & $\surd$ & NA & NA  & NA  & NA &
 	$\surd$ & NA & $\surd$  \\
\midrule
\citet{HonkiszK018} &
 	$\surd$ & $\surd$ & $\surd$ & $\surd$ & $\surd$ & NA &
 	$\surd$ & $\surd$ &$\surd$  \\
\midrule
\citet{derAaDiCiccio19} &  
 	$\surd$ & $\surd$ & $\surd$ & - & - & - &
 	$\surd$ & $\surd$ & --  \\
\midrule  
\citet{DBLP:conf/caise/0003W0LLZLW20} & 
 	$\surd$ & $\surd$ & $\surd$ & - & - & - &
 	NA & NA & NA \\
    \bottomrule
    \end{tabular}
    }
\end{table}
%\end{adjustbox}

% During the recording of PET elements, we discover some disagreements between what is reported in the paper and what the tool really extract.
% For instance,  the tool proposed by \citet{GoncalvesSB11} does not extract any gateway element and its relations; the tool proposed by \cite{EpureMHDS15} extracts neither actor and actor performer relation nor gateway and its components (condition specification) and its relations; and the tool proposed by \citet{DBLP:conf/caise/0003W0LLZLW20} does not extract any relation at all.
% In addition, the opposite situation also happened.
% Indeed, we discover that it is possible to extract the actor and its relation in the tool proposed by \citet{derAaDiCiccio19} and by \citet{HonkiszK018}.
% Therefore, we recorded that information in our evaluation.

As described above, we adapted the source code of the available tool implementations in order to identify the PET entities and relations that the approaches extract when analyzing textual documents, thus allowing us to obtain standardized output for the \tools different approaches. We report the details of these source code adaptations together with all the material to reproduce the experiments in the paper's repository\footnote{Available at \url{github.com/patriziobellan86/ProcessExtractionFromTextSotaAndChallenges}.}.
%Once identified the elements to extract, we extract them. In this step we recorded the extraction of process model elements and their relations directly in the tool's source code when a tool analyses the \petonly text documents.
%Recording elements and relations in the source code of tools was necessary since in our quantitative evaluation we are interested in comparing the ability of the tools to extract process model elements and relations directly from the the raw textual descriptions of the dataset as explained at the beginning of the Section in order to test all the tools on a common task. 
%The details regarding where we recorded the entities and relations extracted in the tool code are reported in the paper repository.

%In the third step of our evaluation pipeline, we  compute the scores based on the metrics illustrated in the next Section and produced the results that we illustrate in Section~\ref{ssec:results-relaxed-comparison}. 

% we already reported the repository url in previous section at \\
%\gitrepo.

% \btext{
% Finally, in the fourth step of our evaluation pipeline, we quantitatively compared the performance of the tools.
% In this evaluation, we have decided to compare the \tools tools in two groups of extraction tasks: first, the extraction of the PET process model elements Activity, Activity Data, Actor, XOR Gateway, AND Gateway, and Condition Specification; second, the extraction of the PET process model relations Uses, Actor Performer, and Flow.
% }
 
\subsection{Evaluation Metrics}
\label{ssec:quantitative-metrics}

We quantify the performance of the different tools by comparing the entities and relations that they extract to the ones in the gold standard, where we assign one of three options for each entity or relation:
%We defined the extraction of elements or element relations from texts as a \textit{classification problem}.
%We define three possible classification outcomes: 
(i) \textit{true positive} if the entity or relation is correctly extracted from the text and is present in the gold-standard dataset, (ii) \textit{false positive} if the extracted entity or relation is not present in the gold-standard dataset, and (iii) \textit{false negative} if an entity or relation is present in the gold-standard dataset but it is not extracted by a tool.
%At the end of the extraction, we calculate the sum of True~positive, False~positive, and False~negative.

Note that in information extraction, when comparing tools on a common dataset, an extracted entity is considered a true positive if and only if the span of words of the entity extracted strictly corresponds to the span of words of the gold-standard annotation, and an extracted relation is considered a true positive if and only if the span of words of the source and of the target elements of the relation are classified as true positive and the relation label predicted corresponds to the gold-standard relation label. 
However, the tools that we  compared were not originally developed to extract the process entities and their relations as defined in the \PETD. Therefore, rather than, for example, extracting ``the office'' as an actor from the sentence \emph{the office sends the email}), a tool may just return ``office''. 
%. Therefore, performing a strict comparison may hide their real extraction performance.
%For example, it may be the case that a tool correctly extracts the core part of an element (e.g., it correctly extracts ``office'' as \textit{actor} in \emph{the office sends the email}), but it misses the article.
To not punish tools for this, we relax the comparisons of the predicted span of words of elements with a window of 1 word, either on the left or on the right of the prediction border, thus still yielding a true positive when extracting ``office'' rather than ``the office''.\footnote{For the sake of completeness we did perform also a strict evaluation of the tools. The interested reader can find these results in the paper repository.}

Given these assessments per entity and relation, we use precision, recall, and F-1 score for each element and each relation class, to compare the extraction performance of the tools. 
\\
\noindent\textbf{Precision} is the fraction of elements or relations correctly extracted by a tool among all retrieved instances.
We calculate precision as:
\begin{align*}
precision =  \frac{True~positive}{True~positive + False~positive}
\end{align*}
%\\
\noindent\textbf{Recall} is the fraction of the number of positive cases a tool correctly extracts from texts, over all the cases in the dataset. 
We calculate recall as:
\begin{align*}
recall =  \frac{True~positive}{True~positive + False~negative}
\end{align*}
%\\
\noindent\textbf{F-1 score} is the harmonic mean of the precision and recall.
We calculate the F-1 score as:
\begin{align*}
F1~score = 2\times\frac{Precision~\times~Recall}{Precision~+~Recall}
\end{align*}

\subsection{Tool Comparison}
\label{ssec:results-relaxed-comparison}
%

%%%%%%%%
In this section, we provide the results of the tool comparison, of which we summarize the main statistics in Table~\ref{tab:relax-match-elements}. Here we present the micro average results across the different texts in the PET dataset, allowing us to account for differences in size and coverage.

% of the tools when we relax the comparison of process elements extracted with a window of one word either on the left or on the right border of the span of words prediction.
%We report the results of in Table~\ref{tab:relax-match-elements} grouped by Entity (Table~\ref{tab:entities}) and Relation (Table~\ref{tab:relations}). 

\begin{table}[!htbp]
	\caption{Micro-averaged results of the quantitative comparison.}
	\label{tab:relax-match-elements}
	\begin{subtable}[t]{\textwidth}
		\centering
		\resizebox{\textwidth}{!}{\begin{tabular}{lcccccccccccccccccc} 	
				\toprule
				\multirow{3}{*}{\textbf{Paper}} & 
				\multicolumn{3}{c}{\multirow{2}{*}{\textbf{Activity}}} & 
				\multicolumn{3}{c}{\textbf{\emph{Activity}}} & 
				\multicolumn{3}{c}{\multirow{2}{*}{\textbf{\emph{Actor}}}} &
				\multicolumn{3}{c}{\textbf{\emph{XOR}}} & 
				\multicolumn{3}{c}{\textbf{\emph{Condition}}} & 
				\multicolumn{3}{c}{\textbf{\emph{AND}}} \\
				&  & & & \multicolumn{3}{c}{\textbf{\emph{Data}}} & & & &  
				\multicolumn{3}{c}{\textbf{\emph{Gateway}}} & 
				\multicolumn{3}{c}{\textbf{\emph{Specification}}} & 
				\multicolumn{3}{c}{\textbf{\emph{Gateway}}} \\
				
				\cmidrule(lr){2-4} 
				\cmidrule(lr){5-7} 
				\cmidrule(lr){8-10}  
				\cmidrule(lr){11-13} 
				\cmidrule(lr){14-16} 
				\cmidrule(lr){17-19} 
				&  pr & rec & f1 & pr & rec & f1 & pr & rec & f1 & pr & rec & f1 & pr & rec & f1 & pr & rec & f1 \\ 
				%& pr & rec & f1 \\
				%  
				\midrule 
				\citet{GoncalvesSB11} &  
				%% Activity
				0.53 & 0.42 & 0.45 &
				%% Activity Data
				0.34 & 0.29 & 0.29 &
				%% Actor
				0.62 & 0.32 & 0.41 &
				% Further Spec.
				%- & - & - &
				% XOR
				- & - & - &
				% Condition Specificaiton
				- & - & - &
				% AND
				- & - & - \\
				\midrule  
				\cite{Friedrich11} & 
				%% Activity 
				0.59 & \textbf{0.82} & 0.68 & 
				%% Activity Data
				\textbf{0.40} & \textbf{0.52} & \textbf{0.44} &
				%% Actor
				\textbf{0.69} & \textbf{0.66} & \textbf{0.65} &
				%% Further Specification
				%\textbf{0.17} & \textbf{0.27} & \textbf{0.19} & 
				%% XOR
				0.50 & 0.26 & 0.32 & 
				%% Condition Specification
				- & - & - &
				%% And Gateway
				\textbf{0.12} & \textbf{0.12} & \textbf{0.12} \\
				\midrule
				\cite{EpureMHDS15} & 
				% Activity	
				0.13 & 0.15 & 0.13 &	
				% Activity Data  
				0.26 & 0.34 & 0.27 &
				% Actor
				- & - & - &	
				% Further
				%- & - & - &
				% XOR
				- & - & - &
				%. Condition Specification
				- & - & - &
				%. AND
				- & - & - \\
				\midrule
				\citet{HonkiszK018} &
				% Activity
				\textbf{0.70} & 0.72 & \textbf{0.70} &
				% Activity Data
				0.21 & 0.25 & 0.22 &
				% Actor
				0.52 & 0.44 & 0.44 &
				% Further Spec.
				%- & - & - &
				% XOR Gateway
				\textbf{0.66} & \textbf{0.39} & \textbf{0.47} & 
				% Condition Specification
				\textbf{0.35} & \textbf{0.25} & \textbf{0.28} &
				% AND 
				- & - & - \\
				\midrule  
				\citet{derAaDiCiccio19} &  
				% Activity
				0.08 & 0.14 & 0.10 & 
				% Activity Data
				0.00 & 0.00 & 0.00 &
				% Actor
				0.65 & 0.41 & 0.48 &
				% Further Spec
				%- & - & - &
				% Xor
				- & - & - &
				% Condition Specification
				- & - & - &
				% AND 
				- & - & - \\
				\midrule
				\citet{DBLP:conf/caise/0003W0LLZLW20} & 
				% Activity
				0.07 & 0.02 & 0.03 &
				% Activity Data
				0.04 & 0.09 & 0.06 &
				% Actor
				0.47 & 0.33 & 0.37 &
				% Further
				%- & - & - &
				% XOR
				- & - & - &
				% Cond. Spec.
				- & - & - &
				%  AND
				- & - & - \\
				\bottomrule
			\end{tabular}
		}
		\caption{Process Entities.}
		\label{tab:entities}
	\end{subtable}\\
	\begin{subtable}[t]{\textwidth}	
		\centering
		\resizebox{.65\textwidth}{!}{\begin{tabular}{lccccccccc} 
				
				\toprule
				\multirow{3}{*}{\textbf{Paper}} & 
				\multicolumn{3}{c}{\multirow{2}{*}{\textbf{\emph{Uses}}}} & 
				\multicolumn{3}{c}{\textbf{\emph{Actor}}} & 
				\multicolumn{3}{c}{\multirow{2}{*}{\textbf{\emph{Flow}}}} \\
				&
				&
				&
				&
				\multicolumn{3}{c}{\textbf{\emph{Performer}}} & 
				&
				&
				\\
				\cmidrule(lr){2-4} \cmidrule(lr){5-7} \cmidrule(lr){8-10}
				&  
				pr & rec & f1 &
				pr & rec & f1 &
				pr & rec & f1 \\
				\midrule  
				\citet{GoncalvesSB11} & 
				% Uses
				0.87 & 0.23 & 0.35 &
				% Actor-Performer
				0.85 & 0.33 & 0.44 &
				% Actor Recipient
				% - & - & - & 	
				% Further Spec. 
				%- & - & - &
				% Flow
				0.64 & 0.07 & 0.12 \\
				% Same Gateway	
				%- & - & - \\
				%  	
				\midrule  
				\cite{Friedrich11} & 
				% Uses
				\textbf{0.97} & \textbf{0.47} & \textbf{0.61} &
				% Actor-Performer
				\textbf{0.99} & \textbf{0.75} & \textbf{0.84} &
				% Actor Recipient
				% - & - & - & 	
				% Further Spec. 
				%\textbf{0.56} & \textbf{0.27} & \textbf{0.35} &
				% Flow
				0.92 & 0.27 & 0.39 \\
				% Same Gateway	
				%- & - & - \\
				%  		
				\midrule
				\cite{EpureMHDS15} & 
				% Uses
				0.41 & 0.06 & 0.10 &
				% Actor-Performer
				- & - & - &
				% Actor Recipient
				% - & - & - & 	
				% Further Spec. 
				%- & - & - &
				% Flow
				0.23 & 0.01 & 0.03 \\
				% Same Gateway	
				%- & - & - \\
				%  	
				\midrule
				\citet{HonkiszK018} & 
				% Uses
				0.82 & 0.19 & 0.29 &
				% Actor-Performer
				0.92 & 0.51 & 0.61 &
				% Actor Recipient
				% - & - & - & 	
				% Further Spec. 
				%- & - & - &
				% Flow
				\textbf{0.97} & \textbf{0.31} & \textbf{0.44} \\
				% Same Gateway	
				%\textbf{0.07} & \textbf{0.07} & \textbf{0.07} \\
				%  	
				\midrule 
				\citet{derAaDiCiccio19} & 
				% Uses
				0.00 & 0.00 & 0.00 &
				% Actor Performer
				0.34 & 0.07 & 0.11 &
				% Actor Recipient
				% - & - & - & 	
				% Further Spec. 
				%- & - & - &
				% Flow
				- & - & - \\
				% Same Gateway	
				%- & - & - \\
				%
				\midrule
				\citet{DBLP:conf/caise/0003W0LLZLW20} & 
				% Uses
				- & - & - &
				% Actor-Performer
				- & - & - &
				% Actor Recipient
				% - & - & - & 	
				% Further Spec. 
				%- & - & - &
				% Flow
				- & - & - \\
				% Same Gateway	
				%- & - & - \\
				%  	
				\bottomrule
			\end{tabular}
		}
		\caption{Process Relations.}
		\label{tab:relations}
	\end{subtable}    
	
\end{table}

We start with the analysis of activities, as they are central to any process.
If we focus only on the \emph{Activity} entity in PET (that is the action within an activity), then the 
best performance in terms of precision and f1-score is achieved by \citet{HonkiszK018} (0.70 precision and f1), whereas the tool by \citet{Friedrich11} achieves the highest recall, with 0.82.
When also considering the \emph{Activity Data} entity, e.g., the business objects, and the corresponding Uses relation that associates the data entity to actions, then the tool of \citet{Friedrich11} achieves the overall best performance.

On the three elements \emph{Activity}, \emph{Uses}, and \emph{Activity Data}, together the tools of \citet{GoncalvesSB11} and \citet{HonkiszK018} show overall an intermediate performance, followed by the tool of \citet{EpureMHDS15}, and then those of \citet{derAaDiCiccio19, DBLP:conf/caise/0003W0LLZLW20}, which show particularly low performances on all these three elements as defined in the \PETD. If we consider \emph{Activity}, \emph{Uses}, and \emph{Activity Data} separately we can notice a drastic drop in performance, for the three best tools \citep{Friedrich11,GoncalvesSB11,HonkiszK018}, when they focus on \emph{Activity Data}, with respect to \emph{Activity}. Indeed we can safely say that for this entity none of the tools perform in a satisfactory manner. A further interesting insight on the three top scorers is the difference between precision and recall for the \emph{Uses} relation. Here all the tools have very high precision but fail in reaching a satisfactory recall.

% The tools of \citet{derAaDiCiccio19, EpureMHDS15, DBLP:conf/caise/0003W0LLZLW20} have very low scores that provide their inability to extract \textit{activity} as defined in the PET annotation schema.
% Also, the object an activity acts on is extracted by all the tools.
% Different from the situation depicted for activity extraction, the tool of \citet{HonkiszK018} shows severe difficulties to extract this element while the tool of \citet{Friedrich11} shows the best scores in all the three metrics.
% The tool of \citet{GoncalvesSB11} is the second-best scoring tool.
% But, overall the scores for the extraction of this element are not satisfactory.
% The tools of \citet{derAaDiCiccio19, EpureMHDS15, DBLP:conf/caise/0003W0LLZLW20} show very low scores. These demonstrate their inability to extract this element from texts.
% The results of the \textit{uses} relation in Table \ref{tab:relax-match-relations} show the same situation for the extraction of \textit{activity data} element.
% The tool of  \citet{Friedrich11} provides the highest scores in all three metrics. The tool of \citet{GoncalvesSB11} is the second-best scoring tool with good precision but low recall.
% The other tools are not able to extract this element satisfactorily.

When we focus on the extraction of \textit{Actor} entities and \textit{Actor performer} relations, we can observe a different situation. In fact, while the best scorer remains the tool of \citet{Friedrich11}, which is the only one having satisfactory results for both elements, we can notice a more homogeneous situation with no tool having excellent performances on both elements but also no tool having dramatic low results. The only exception here is the tool of \citet{derAaDiCiccio19} on \emph{Actor Performer}, which is likely due to the extremely low performance of this tool on the \emph{Activity} entity. 

% from text is performed in all the tools, but \citet{EpureMHDS15}.
% The scores show that the most precise tools are the proposal of \citet{derAaDiCiccio19}, \citet{Friedrich11}, and \citet{GoncalvesSB11}.
% The tool of  \citet{Friedrich11} reached the highest scores in all the metrics, while the tools of \citet{derAaDiCiccio19, GoncalvesSB11} show some difficulties in recalling all the actors mentioned in texts.
% The tool of \citet{DBLP:conf/caise/0003W0LLZLW20} is not able to provide satisfactory results, overall.
% The \textit{actor performer} relation is extracted by \citet{derAaDiCiccio19}, \citet{Friedrich11}, \citet{GoncalvesSB11}, and \citet{HonkiszK018}.
% The scores reported in Table \ref{tab:relax-match-relations} show that the tool of \citet{Friedrich11} is the most performing tool.
% It can correctly extract most of this relation from texts.
% Indeed, it reached an almost perfect precision and a good recall.
% In general, this tool provided satisfactory results for the extraction of the actors responsible for activities.

The other element extracted by most tools is the \textit{Flow} relation among behavioral entities.
All tools here have a much better precision than recall. This latter one is not fully satisfactory for all the tools considered. 
The best tool on the \textit{Flow} relation is the one of \citet{HonkiszK018} that has the highest scores in all the three metrics followed by \citet{Friedrich11}. Interestingly enough, the tool of \citet{Friedrich11} extracts slightly more activities from texts than the one of \citet{HonkiszK018}, but it is less efficient than \citet{HonkiszK018} in extracting XOR gateways (see below). Since the gateways are also elements of the control flow, this may be a possible cause of the slightly  best performance of the tool of \citet{HonkiszK018} on the \textit{Flow} relation. 

Finally, gateways are extracted only by the tools of \citet{Friedrich11} and \citet{HonkiszK018}.
The one of \citet{HonkiszK018} performed slightly better on \textit{XOR Gateway} in all three metrics.
The \textit{Condition Specification} entity is extracted by the tool of \citet{HonkiszK018} only, with rather low scores. Similarly, the \textit{AND Gateway} element is extracted by \citet{Friedrich11} only, with extremely low scores. 

Analyzing the overall performance of the tools, it is possible to differentiate three groups: satisfactory performance, medium performance and low performance. 
In the first group there are the contributions of \citet{Friedrich11} and \citet{HonkiszK018}. They propose tools that show satisfactory performances on a reasonable set of elements which comprise activities, actors and the control-flow structure.
In particular, the tool of \citet{Friedrich11} shows the overall best performances for what concerns the extraction of activity (and its components) and actor from texts. while the tool of \citet{HonkiszK018} shows the overall best performance for the extraction of the control flow structure (flow and gateways). 
The second group is composed by the tool of \citet{GoncalvesSB11}. This tool is not able to address gateways and has an overall lower performance on the remaining elements w.r.t. the first two tools. 
Finally, the remaining three tools do not produce satisfactory results on the \PETD. A partial explanation for tool of \citet{derAaDiCiccio19} may be related to the fact that they target the extraction of Declarative relations, even though the higher performance on \emph{Actor} w.r.t. \emph{Activity} of this tool remains rather surprising.  
The low performance of the tools of~\citet{EpureMHDS15} and \citet{DBLP:conf/caise/0003W0LLZLW20} is likely due to their focus on a specific subset of process-related descriptions, respectively targeting the archaeological domain and collections of manuals and recipes. The results obtained in our study indicate that these tools generalize poorly to other types of text.

%can be explained by the fact that they target descriptions of process executions (in the archaeological domain) and collections of manuals and recipes, respectively. Our evaluation shows that focusing on a text that is related, but different, different from the description of a process models, is likely not a good strategy here, as the models are not able to scale to this type of text. 

As a limitation of our study we need to observe that our evaluation was performed on the \PETD, which is an annotated version of the dataset proposed originally by \citet{Friedrich11}. One may argue that this may have favored that tool w.r.t. others. Nonetheless we need to note that (i) other tools that used subsets of the dataset by \citet{Friedrich11} (see Table~\ref{tab:experimental-evaluation})
 did not have particularly high performances, and (ii) the paper of \citet{HonkiszK018} does not report any evaluation of their tool (see Section~\ref{ssec:experimental-evaluation}) and nonetheless was one with satisfactory scores. While an enlargement of the \PETD is recommendable (see Section~\ref{sec:discussion}), this highlights the value and the potential of the evaluation pipeline proposed in this paper.

 %!TEX root = .././main.tex
\section{Limitations and Challenges for \PET }
\label{sec:discussion}
     
The analysis proposed in the previous sections reveals several problematic aspects of state-of-the-art contributions. % in \pet. 
In this section, we elaborate on four sets of limitations concerning
\begin{enumerate*}[(i)]
	\item the data(sets);
	\item the techniques adopted so far;
	\item the conducted experimental evaluation; and 
	\item the pipeline proposed.
\end{enumerate*} that emerge from the qualitative and quantitative evaluation and highlight challenges that the community may need to address in the future to produce significant advances in this area.  

 %%%%%% DATA %%%%%%%%%%%%%%% 
% \paragraph*{The data.}
% \label{ssec:the-data}

A crucial set of limitations concern the \textbf{datasets} used in the primary studies, which are problematic in different ways: first, there is a lack of publicly available annotated data. In fact most of the datasets reduce to subsets of the dataset originally proposed by \citet{Friedrich11}. The only other dataset publicly available is the one of \citet{DBLP:conf/caise/0003W0LLZLW20}. Nonetheless it consists of instruction manuals and food recipes, whose complexity in terms of control flow is often simpler than the one of business or work processes. This critical aspect is confirmed also by our quantitative evaluation in which the tool is unable to show acceptable performances. Despite of its importance, the 47 texts proposed by \citet{Friedrich11} can be hardly considered as representative of the variety of process descriptions and process models one may have in real scenarios and are not enough to support the training and testing required by many sophisticated machine-learning techniques. Thus, the first challenge for the community would be to enlarge the amount of textual data available, so as to cover different phenomena. This may mimic what the community of Process Mining has done with event log datasets (e.g., via the BPI challenges) and could prove extremely beneficial. 
Second, the dataset of \citet{Friedrich11}, represents pairs of process descriptions and process models, without any intermediate form of annotation. As such, the dataset was not ready to be used to comparatively evaluate the extraction of specific elements, such as activities, actors, gateways, events, and so on. In our opinion, this characteristics has had a negative impact on the development of \pet as a solid field, with reference common challenges on well defined datasets/benchmarks and metrics. This has likely originated the heterogeneous situation depicted in Table \ref{tab:experimental-evaluation}, where the different contributions can be hardly compared on common ground, instead of creating a community focused on reference tasks, possibly of growing difficulty, to be dealt with in a competitive manner in a way similar to what happens in e.g., the NLP or Vision communities. The \PETD partly solves this problem but it should be extended both in terms of number of texts and in annotation tags. Thus a challenge for the community to address should be: (i) to extend this dataset, and (ii) to use it e.g., to propose competitions on specific tasks related to \pet, following the  example of the quantitative comparison of the tools proposed in Section~\ref{sec:quantitative-comparison}.

% %%%%%% TECHNIQUES %%%%%%%%%%%%%%%
% \paragraph{The techniques.}
% \label{ssec:the-techniques}

The limited availability of data has had a direct impact on the \textbf{techniques} that have been so far applied to tackle \pet. In particular, it has made the usage of neural models virtually impossible, due to their dependency on large amounts of training data.
Indeed, most of the approaches present in literature and examined in this paper resort to using rule-based strategies, which often suffer the problem of generality, as they are usually tailored to specific scenarios, writing styles, and formatting of documents, despite the usage of external resources such as WordNet, and VerbNet to increase the coverage and robustness of the rules themselves. 
A further intriguing aspect of state-of-the-art contributions is that most works extract syntactic and semantic information through the analysis of the plain text, whereas only a few papers provide a direct semantic representation of the text using semantic word embeddings. Given the impact that semantic word embeddings have recently had on different tasks in NLP, and the importance of it enhancing, e.g., the performance of statistical classifiers, this also opens interesting questions for future research efforts. 
Similarly, the usage of external resources and NLP techniques/tools (such as coreference resolution tools) is almost absent in all the works we have examined.   
Given the recent advances in NLP, this observation raises an interesting question for the community: 
\textit{Can we increase the frameworks' performance by leveraging recent NLP advancements and up-to-date resources?} This question becomes even more important in the Large Language Models era, which may present further opportunities for the extraction of procedural information from texts.

As already discussed in Section~\ref{sec:qualitative-comparison} a further important limitation concerns the \textbf{experimental evaluation} conducted in the primary studies themselves. In particular, we observed the absence of (i) common tasks which could foster a  comparative development of techniques and tools and (ii) commonly agreed metrics to be used for such tasks. To overcome these issues, this paper proposes a first set of benchmarking tasks. They, nonetheless should be expanded with the extraction of other elements and also the addition of specific tasks related to the composition of a process model. Furthermore, the community should make a serious effort to open a discussion on the metrics to be used. By looking at Table \ref{tab:experimental-evaluation}, we can see that most contributions analyzed in this paper either rely on information retrieval measures or on graph-based measures. A clear discussion on which metrics to use and for which tasks is therefore needed. 
Another problem that deserves consideration is the fact that these metrics do not distinguish between possible kinds of errors that can be generated. For example, would it be better to identify a gateway but assign it to the wrong type, or to not identify the gateway at all? Would it be better to associate an activity with the wrong actor or to just extract the activity? Answering these questions is a challenging task, as extracting more information may help the manual refinement of the process, but errors that are not rectified may have significant negative consequences on the process model. 
A further problem is given by the fact that the metrics proposed in literature mainly consider the graph structure of the process model and do not make a distinction between equivalent (very similar) process model fragments. As a simple example, how should we evaluate when two activities are represented as a single, compound activity in the diagram? Different choices could be made, but it would be good for the community to reflect on them and make some clear proposal of tasks and related measures.

% %%%%%% PIPELINE %%%%%%%%%%%%%%%
% \paragraph{The pipeline.}
% \label{ssec:the-pipeline}

Concerning the overall transformation \textbf{pipeline} from text to process model, the analysis of the state-of-the-art contributions shows efforts in both the direct and the two-step approaches. While the direct approach may appear more attractive due to the popularity of end-to-end machine learning approaches, the big question here is \textit{is there a way to collect the massive set of data that would enable the construction of a generic direct pipeline?}  
The results of the comparison of the tools show that two-step approaches outperformed one-step approaches by a large margin. Also, a two-step approach would allow an overall pipeline to be decomposed into different (sub-)tasks, for which individual techniques can be developed.
As an example, we have noticed that the task of \textit{filtering out uninformative information} is rarely considered in state-of-the-art contributions but may be extremely important in real-world procedural documents. This could be because there are almost no uninformative textual fragments in the reference dataset of \citet{Friedrich11} even though the pipeline should drive the construction of the datasets and not the other way around. Another challenge concerns the ordering of the different sub-tasks and the impact it can have on the overall pipeline. For example, do we need to resolve co-references before filtering out uninformative sentences from a text or can we postpone this task until later?

Further questions concern the usage of the intermediate representation. First of all, no clear reasons are given in the different papers on why a specific world model was used, and no study exists to assess whether some are better and offer a more robust intermediate representation. 
Second, sharing the data contained in different world models built by different \pet frameworks is currently very hard. This is extremely problematic, and having a common world model (or at least a way to move data across different ones) would boost a real modularization of $\fa$ and $\fb$, allowing reuse and the investigation of single components of the pipeline one at a time. Knowledge graphs could offer a framework for an intermediate representation, but again, this should be a decision for the entire community to make.
Finally, an important benefit of splitting the transformation pipeline into smaller tasks is that approaches can build on each other, e.g., one paper may produce a highly accurate approach for activity extraction, whereas work by author authors takes these activities and focuses on the detection of control-flow relations between them. In the current state of the field, such cumulative progress is lacking, since each of the primary studies starts from scratch.

Last but not least, we expect the proposal of new pipelines based on the usage of pre-trained Large Language Models (GPT3, for example). In fact, the usage of these emerging techniques may partly solve the lack of training data emphasized at the beginning of this section. While this may be a good news for this field, even if the performance of these models on this specific domain remain to be assessed, this makes the definition of benchmarking tasks, texts, and metrics even more compelling. Otherwise the risk is a further fragmentation of the field in a plethora of non-comparable works, and a difficulty to further build on previous works.

\section*{Conclusions}
\label{sec:conclusions}
In this paper we provide a qualitative analyses of \papers state-of-the-art contributions on \pet, published specifically in the area of Business Process Management from 2010, and a quantitative analyses of \tools working tools publicly available from the \papers paper, to better understand their contributions and limitations and to identify which are important challenges in this topic the research community should address. 

The qualitative analyses show an heterogeneity of techniques, elements extracted and evaluation conducted, that are often impossible to compare in a direct manner. To overcome this difficulty we propose a quantitative comparison of the tools proposed by the different papers on the unifying task of process model entity and relation extraction so as to be able to compare them directly. The results show three distinct groups of tools in terms of performance, with no tool obtaining very good scores and also serious limitations in terms of elements extracted. Moreover, the proposed evaluation pipeline can be considered a reference task on a well defined dataset and metrics that can be used to compare new tools in a way similar to what happens in e.g., the natural language processing or computer vision communities. 
We conclude the paper with a reflection on the results of the qualitative and quantitative evaluation, and in particular on the limitations and challenges on the data(sets); the techniques adopted so far; the experimental evaluations conducted; and the pipelines proposed. These limitations and related challenges can offer a stimulus for the community to identify reference tasks (e.g., by exploiting an idea similar to what happened with the BPI challenges\footnote{See \url{https://www.tf-pm.org/competitions-awards/bpi-challenge}}) that the community may need to address in the future in a rigorous comparative manner to produce significant advances in this area. 

\setstretch{2.0}
\bibliographystyle{apacite}
\bibliography{biblio}

\begin{thebibliography}{}

\bibitem [\protect \citeauthoryear {%
Achour%
}{%
Achour%
}{%
{\protect \APACyear {1998}}%
}]{%
Achour98}
\APACinsertmetastar {%
Achour98}%
\begin{APACrefauthors}%
Achour, C\BPBI B.%
\end{APACrefauthors}%
\unskip\
\newblock
\APACrefYearMonthDay{1998}{}{}.
\newblock
{\BBOQ}\APACrefatitle {{Guiding Scenario Authoring}} {{Guiding Scenario Authoring}}.{\BBCQ}
\newblock
\BIn{} \APACrefbtitle {{Information Modelling and Knowledge Bases {X:} 8th European-Japanese Conferences on Information Modelling and Knowledge Bases}} {{Information Modelling and Knowledge Bases {X:} 8th European-Japanese Conferences on Information Modelling and Knowledge Bases}}\ (\BVOL~51, \BPGS\ 152--171).
\newblock
\APACaddressPublisher{}{{IOS} Press}.
\PrintBackRefs{\CurrentBib}

\bibitem [\protect \citeauthoryear {%
Ackermann%
, Neuberger%
\BCBL {}\ \BBA {} Jablonski%
}{%
Ackermann%
\ \protect \BOthers {.}}{%
{\protect \APACyear {2021}}%
}]{%
DBLP:conf/caise/AckermannNJ21}
\APACinsertmetastar {%
DBLP:conf/caise/AckermannNJ21}%
\begin{APACrefauthors}%
Ackermann, L.%
, Neuberger, J.%
\BCBL {}\ \BBA {} Jablonski, S.%
\end{APACrefauthors}%
\unskip\
\newblock
\APACrefYearMonthDay{2021}{}{}.
\newblock
{\BBOQ}\APACrefatitle {Data-Driven Annotation of Textual Process Descriptions Based on Formal Meaning Representations} {Data-driven annotation of textual process descriptions based on formal meaning representations}.{\BBCQ}
\newblock
\BIn{} \APACrefbtitle {{Advanced Information Systems Engineering - 33rd International Conference, CAiSE 2021, Proceedings}} {{Advanced Information Systems Engineering - 33rd International Conference, CAiSE 2021, Proceedings}}\ (\BVOL\ 12751, \BPGS\ 75--90).
\newblock
\APACaddressPublisher{}{Springer}.
\PrintBackRefs{\CurrentBib}

\bibitem [\protect \citeauthoryear {%
Adamo%
\ \protect \BOthers {.}}{%
Adamo%
\ \protect \BOthers {.}}{%
{\protect \APACyear {2017}}%
}]{%
DBLP:conf/aiia/AdamoBFGGS17}
\APACinsertmetastar {%
DBLP:conf/aiia/AdamoBFGGS17}%
\begin{APACrefauthors}%
Adamo, G.%
, Borgo, S.%
, {Di Francescomarino}, C.%
, Ghidini, C.%
, Guarino, N.%
\BCBL {}\ \BBA {} Sanfilippo, E\BPBI M.%
\end{APACrefauthors}%
\unskip\
\newblock
\APACrefYearMonthDay{2017}{}{}.
\newblock
{\BBOQ}\APACrefatitle {{Business Processes and Their Participants: An Ontological Perspective}} {{Business Processes and Their Participants: An Ontological Perspective}}.{\BBCQ}
\newblock
\BIn{} \APACrefbtitle {{AI*IA 2017 Advances in Artificial Intelligence - XVIth International Conference of the Italian Association for Artificial Intelligence, Proceedings}} {{AI*IA 2017 Advances in Artificial Intelligence - XVIth International Conference of the Italian Association for Artificial Intelligence, Proceedings}}\ (\BVOL\ 10640, \BPGS\ 215--228).
\newblock
\APACaddressPublisher{}{Springer}.
\PrintBackRefs{\CurrentBib}

\bibitem [\protect \citeauthoryear {%
Adamo%
, {Di Francescomarino}%
\BCBL {}\ \BBA {} Ghidini%
}{%
Adamo%
\ \protect \BOthers {.}}{%
{\protect \APACyear {2020}}%
}]{%
DBLP:conf/caise/AdamoFG20}
\APACinsertmetastar {%
DBLP:conf/caise/AdamoFG20}%
\begin{APACrefauthors}%
Adamo, G.%
, {Di Francescomarino}, C.%
\BCBL {}\ \BBA {} Ghidini, C.%
\end{APACrefauthors}%
\unskip\
\newblock
\APACrefYearMonthDay{2020}{}{}.
\newblock
{\BBOQ}\APACrefatitle {{Digging into Business Process Meta-models: {A} First Ontological Analysis}} {{Digging into Business Process Meta-models: {A} First Ontological Analysis}}.{\BBCQ}
\newblock
\BIn{} \APACrefbtitle {{Advanced Information Systems Engineering - 32nd International Conference, CAiSE 2020, Proceedings}} {{Advanced Information Systems Engineering - 32nd International Conference, CAiSE 2020, Proceedings}}\ (\BVOL\ 12127, \BPGS\ 384--400).
\newblock
\APACaddressPublisher{}{Springer}.
\PrintBackRefs{\CurrentBib}

\bibitem [\protect \citeauthoryear {%
Bellan%
, van~der Aa%
, Dragoni%
, Ghidini%
\BCBL {}\ \BBA {} Ponzetto%
}{%
Bellan%
\ \protect \BOthers {.}}{%
{\protect \APACyear {2022}}%
}]{%
DBLP:conf/bpm/BellanADGP22}
\APACinsertmetastar {%
DBLP:conf/bpm/BellanADGP22}%
\begin{APACrefauthors}%
Bellan, P.%
, van~der Aa, H.%
, Dragoni, M.%
, Ghidini, C.%
\BCBL {}\ \BBA {} Ponzetto, S\BPBI P.%
\end{APACrefauthors}%
\unskip\
\newblock
\APACrefYearMonthDay{2022}{}{}.
\newblock
{\BBOQ}\APACrefatitle {{PET:} An Annotated Dataset for Process Extraction from Natural Language Text Tasks} {{PET:} an annotated dataset for process extraction from natural language text tasks}.{\BBCQ}
\newblock
\BIn{} \APACrefbtitle {{Business Process Management Workshops - {BPM} 2022 International Workshops, Revised Selected Papers}} {{Business Process Management Workshops - {BPM} 2022 International Workshops, Revised Selected Papers}}\ (\BVOL~460, \BPGS\ 315--321).
\newblock
\APACaddressPublisher{}{Springer}.
\PrintBackRefs{\CurrentBib}

\bibitem [\protect \citeauthoryear {%
de Almeida~Bordignon%
\ \protect \BOthers {.}}{%
de Almeida~Bordignon%
\ \protect \BOthers {.}}{%
{\protect \APACyear {2018}}%
}]{%
Bordignon2018NaturalLP}
\APACinsertmetastar {%
Bordignon2018NaturalLP}%
\begin{APACrefauthors}%
de Almeida~Bordignon, A\BPBI C.%
, Thom, L\BPBI H.%
, Silva, T\BPBI S.%
, Dani, V\BPBI S.%
, Fantinato, M.%
\BCBL {}\ \BBA {} Ferreira, R\BPBI C\BPBI B.%
\end{APACrefauthors}%
\unskip\
\newblock
\APACrefYearMonthDay{2018}{}{}.
\newblock
{\BBOQ}\APACrefatitle {Natural Language Processing in Business Process Identification and Modeling: {A} Systematic Literature Review} {Natural language processing in business process identification and modeling: {A} systematic literature review}.{\BBCQ}
\newblock
\BIn{} C.~Boscarioli, C\BPBI A.~Costa, S.~de~Avila~e Silva\BCBL {}\ \BBA {} D\BPBI L.~Notari\ (\BEDS), \APACrefbtitle {Proceedings of the {XIV} Brazilian Symposium on Information Systems, {SBSI} 2018} {Proceedings of the {XIV} brazilian symposium on information systems, {SBSI} 2018}\ (\BPGS\ 25:1--25:8).
\newblock
\APACaddressPublisher{}{{ACM}}.
\PrintBackRefs{\CurrentBib}

\bibitem [\protect \citeauthoryear {%
de A.~R.~Gon{\c{c}}alves%
, Santoro%
\BCBL {}\ \BBA {} Bai{\~{a}}o%
}{%
de A.~R.~Gon{\c{c}}alves%
\ \protect \BOthers {.}}{%
{\protect \APACyear {2011}}%
}]{%
GoncalvesSB11}
\APACinsertmetastar {%
GoncalvesSB11}%
\begin{APACrefauthors}%
de A.~R.~Gon{\c{c}}alves, J\BPBI C.%
, Santoro, F\BPBI M.%
\BCBL {}\ \BBA {} Bai{\~{a}}o, F\BPBI A.%
\end{APACrefauthors}%
\unskip\
\newblock
\APACrefYearMonthDay{2011}{}{}.
\newblock
{\BBOQ}\APACrefatitle {{Let Me Tell You a Story - On How to Build Process Models}} {{Let Me Tell You a Story - On How to Build Process Models}}.{\BBCQ}
\newblock
\APACjournalVolNumPages{{J. {UCS}}}{17}{2}{276--295}.
\PrintBackRefs{\CurrentBib}

\bibitem [\protect \citeauthoryear {%
Dumas%
, la Rosa%
, Mendling%
\BCBL {}\ \BBA {} Reijers%
}{%
Dumas%
\ \protect \BOthers {.}}{%
{\protect \APACyear {2013}}%
}]{%
Dumas0031128}
\APACinsertmetastar {%
Dumas0031128}%
\begin{APACrefauthors}%
Dumas, M.%
, la Rosa, M.%
, Mendling, J.%
\BCBL {}\ \BBA {} Reijers, H\BPBI A.%
\end{APACrefauthors}%
\unskip\
\newblock
\APACrefYear{2013}.
\newblock
\APACrefbtitle {Fundamentals of Business Process Management} {Fundamentals of business process management}.
\newblock
\APACaddressPublisher{}{Springer}.
\PrintBackRefs{\CurrentBib}

\bibitem [\protect \citeauthoryear {%
Epure%
, Mart{\'{\i}}n{-}Rodilla%
, Hug%
, Deneck{\`{e}}re%
\BCBL {}\ \BBA {} Salinesi%
}{%
Epure%
\ \protect \BOthers {.}}{%
{\protect \APACyear {2015}}%
}]{%
EpureMHDS15}
\APACinsertmetastar {%
EpureMHDS15}%
\begin{APACrefauthors}%
Epure, E\BPBI V.%
, Mart{\'{\i}}n{-}Rodilla, P.%
, Hug, C.%
, Deneck{\`{e}}re, R.%
\BCBL {}\ \BBA {} Salinesi, C.%
\end{APACrefauthors}%
\unskip\
\newblock
\APACrefYearMonthDay{2015}{}{}.
\newblock
{\BBOQ}\APACrefatitle {Automatic process model discovery from textual methodologies} {Automatic process model discovery from textual methodologies}.{\BBCQ}
\newblock
\BIn{} \APACrefbtitle {{9th {IEEE} International Conference on Research Challenges in Information Science, {RCIS} 2015}} {{9th {IEEE} International Conference on Research Challenges in Information Science, {RCIS} 2015}}\ (\BPGS\ 19--30).
\newblock
\APACaddressPublisher{}{{IEEE}}.
\PrintBackRefs{\CurrentBib}

\bibitem [\protect \citeauthoryear {%
Ferreira%
, Thom%
\BCBL {}\ \BBA {} Fantinato%
}{%
Ferreira%
\ \protect \BOthers {.}}{%
{\protect \APACyear {2017}}%
}]{%
Ferreira17}
\APACinsertmetastar {%
Ferreira17}%
\begin{APACrefauthors}%
Ferreira, R\BPBI C\BPBI B.%
, Thom, L\BPBI H.%
\BCBL {}\ \BBA {} Fantinato, M.%
\end{APACrefauthors}%
\unskip\
\newblock
\APACrefYearMonthDay{2017}{}{}.
\newblock
{\BBOQ}\APACrefatitle {{A Semi-automatic Approach to Identify Business Process Elements in Natural Language Texts}} {{A Semi-automatic Approach to Identify Business Process Elements in Natural Language Texts}}.{\BBCQ}
\newblock
\BIn{} \APACrefbtitle {{{ICEIS} 2017 - Proceedings of the 19th International Conference on Enterprise Information Systems, Volume 3}} {{{ICEIS} 2017 - Proceedings of the 19th International Conference on Enterprise Information Systems, Volume 3}}\ (\BPGS\ 250--261).
\newblock
\APACaddressPublisher{}{SciTePress}.
\PrintBackRefs{\CurrentBib}

\bibitem [\protect \citeauthoryear {%
Ferreres%
, van~der Aa%
, Carmona%
\BCBL {}\ \BBA {} Padr{\'{o}}%
}{%
Ferreres%
\ \protect \BOthers {.}}{%
{\protect \APACyear {2018}}%
}]{%
DBLP:journals/dke/Sanchez-Ferreres18}
\APACinsertmetastar {%
DBLP:journals/dke/Sanchez-Ferreres18}%
\begin{APACrefauthors}%
Ferreres, J\BPBI S.%
, van~der Aa, H.%
, Carmona, J.%
\BCBL {}\ \BBA {} Padr{\'{o}}, L.%
\end{APACrefauthors}%
\unskip\
\newblock
\APACrefYearMonthDay{2018}{}{}.
\newblock
{\BBOQ}\APACrefatitle {Aligning textual and model-based process descriptions} {Aligning textual and model-based process descriptions}.{\BBCQ}
\newblock
\APACjournalVolNumPages{Data Knowl. Eng.}{118}{}{25--40}.
\PrintBackRefs{\CurrentBib}

\bibitem [\protect \citeauthoryear {%
Friedrich%
, Mendling%
\BCBL {}\ \BBA {} Puhlmann%
}{%
Friedrich%
\ \protect \BOthers {.}}{%
{\protect \APACyear {2011}}%
}]{%
Friedrich11}
\APACinsertmetastar {%
Friedrich11}%
\begin{APACrefauthors}%
Friedrich, F.%
, Mendling, J.%
\BCBL {}\ \BBA {} Puhlmann, F.%
\end{APACrefauthors}%
\unskip\
\newblock
\APACrefYearMonthDay{2011}{}{}.
\newblock
{\BBOQ}\APACrefatitle {{Process Model Generation from Natural Language Text}} {{Process Model Generation from Natural Language Text}}.{\BBCQ}
\newblock
\BIn{} \APACrefbtitle {{Advanced Information Systems Engineering - 23rd International Conference, CAiSE 2011. Proceedings}} {{Advanced Information Systems Engineering - 23rd International Conference, CAiSE 2011. Proceedings}}\ (\BVOL\ 6741, \BPGS\ 482--496).
\newblock
\APACaddressPublisher{}{Springer}.
\PrintBackRefs{\CurrentBib}

\bibitem [\protect \citeauthoryear {%
Honkisz%
, Kluza%
\BCBL {}\ \BBA {} Wisniewski%
}{%
Honkisz%
\ \protect \BOthers {.}}{%
{\protect \APACyear {2018}}%
}]{%
HonkiszK018}
\APACinsertmetastar {%
HonkiszK018}%
\begin{APACrefauthors}%
Honkisz, K.%
, Kluza, K.%
\BCBL {}\ \BBA {} Wisniewski, P.%
\end{APACrefauthors}%
\unskip\
\newblock
\APACrefYearMonthDay{2018}{}{}.
\newblock
{\BBOQ}\APACrefatitle {{A Concept for Generating Business Process Models from Natural Language Description}} {{A Concept for Generating Business Process Models from Natural Language Description}}.{\BBCQ}
\newblock
\BIn{} \APACrefbtitle {{Knowledge Science, Engineering and Management - 11th International Conference, {KSEM} 2018, Proceedings, Part {I}}} {{Knowledge Science, Engineering and Management - 11th International Conference, {KSEM} 2018, Proceedings, Part {I}}}\ (\BVOL\ 11061, \BPGS\ 91--103).
\newblock
\APACaddressPublisher{}{Springer}.
\PrintBackRefs{\CurrentBib}

\bibitem [\protect \citeauthoryear {%
Indahyanti%
, Djunaidy%
\BCBL {}\ \BBA {} Siahaan%
}{%
Indahyanti%
\ \protect \BOthers {.}}{%
{\protect \APACyear {2022}}%
}]{%
Indahyanti2022AutoGeneratingBP}
\APACinsertmetastar {%
Indahyanti2022AutoGeneratingBP}%
\begin{APACrefauthors}%
Indahyanti, U.%
, Djunaidy, A.%
\BCBL {}\ \BBA {} Siahaan, D\BPBI O.%
\end{APACrefauthors}%
\unskip\
\newblock
\APACrefYearMonthDay{2022}{}{}.
\newblock
{\BBOQ}\APACrefatitle {Auto-Generating Business Process Model From Heterogeneous Documents: A Comprehensive Literature Survey} {Auto-generating business process model from heterogeneous documents: A comprehensive literature survey}.{\BBCQ}
\newblock
\APACjournalVolNumPages{2022 9th International Conference on Electrical Engineering, Computer Science and Informatics (EECSI)}{}{}{239-243}.
\PrintBackRefs{\CurrentBib}

\bibitem [\protect \citeauthoryear {%
Kitchenham%
\ \BBA {} Charters%
}{%
Kitchenham%
\ \BBA {} Charters%
}{%
{\protect \APACyear {2007}}%
}]{%
Kitchenham07guidelinesfor}
\APACinsertmetastar {%
Kitchenham07guidelinesfor}%
\begin{APACrefauthors}%
Kitchenham, B.%
\BCBT {}\ \BBA {} Charters, S.%
\end{APACrefauthors}%
\unskip\
\newblock
\APACrefYearMonthDay{2007}{}{}.
\newblock
\APACrefbtitle {Guidelines for performing Systematic Literature Reviews in Software Engineering} {Guidelines for performing systematic literature reviews in software engineering}\ \APACbVolEdTR{}{\BTR{}\ \BNUM\ EBSE 2007-001}.
\newblock
\APACaddressInstitution{}{Keele University and Durham University Joint Report}.
\PrintBackRefs{\CurrentBib}

\bibitem [\protect \citeauthoryear {%
L{\'{o}}pez%
, Str{\o}msted%
, Niyodusenga%
\BCBL {}\ \BBA {} Marquard%
}{%
L{\'{o}}pez%
\ \protect \BOthers {.}}{%
{\protect \APACyear {2021}}%
}]{%
DBLP:conf/caise/LopezSNM21}
\APACinsertmetastar {%
DBLP:conf/caise/LopezSNM21}%
\begin{APACrefauthors}%
L{\'{o}}pez, H\BPBI A.%
, Str{\o}msted, R.%
, Niyodusenga, J.%
\BCBL {}\ \BBA {} Marquard, M.%
\end{APACrefauthors}%
\unskip\
\newblock
\APACrefYearMonthDay{2021}{}{}.
\newblock
{\BBOQ}\APACrefatitle {Declarative Process Discovery: Linking Process and Textual Views} {Declarative process discovery: Linking process and textual views}.{\BBCQ}
\newblock
\BIn{} S.~Nurcan\ \BBA {} A.~Korthaus\ (\BEDS), \APACrefbtitle {Intelligent Information Systems - CAiSE Forum 2021, Melbourne, VIC, Australia, June 28 - July 2, 2021, Proceedings} {Intelligent information systems - caise forum 2021, melbourne, vic, australia, june 28 - july 2, 2021, proceedings}\ (\BVOL~424, \BPGS\ 109--117).
\newblock
\APACaddressPublisher{}{Springer}.
\newblock
\begin{APACrefURL} \url{https://doi.org/10.1007/978-3-030-79108-7\_13} \end{APACrefURL}
\newblock
\begin{APACrefDOI} \doi{10.1007/978-3-030-79108-7\_13} \end{APACrefDOI}
\PrintBackRefs{\CurrentBib}

\bibitem [\protect \citeauthoryear {%
Manning%
\ \protect \BOthers {.}}{%
Manning%
\ \protect \BOthers {.}}{%
{\protect \APACyear {2014}}%
}]{%
CoreNLP}
\APACinsertmetastar {%
CoreNLP}%
\begin{APACrefauthors}%
Manning, C\BPBI D.%
, Surdeanu, M.%
, Bauer, J.%
, Finkel, J\BPBI R.%
, Bethard, S.%
\BCBL {}\ \BBA {} McClosky, D.%
\end{APACrefauthors}%
\unskip\
\newblock
\APACrefYearMonthDay{2014}{}{}.
\newblock
{\BBOQ}\APACrefatitle {The Stanford {CoreNLP} Natural Language Processing Toolkit} {The stanford {CoreNLP} natural language processing toolkit}.{\BBCQ}
\newblock
\BIn{} \APACrefbtitle {Proceedings of the 52nd Annual Meeting of the Association for Computational Linguistics, {ACL} 2014, System Demonstrations} {Proceedings of the 52nd annual meeting of the association for computational linguistics, {ACL} 2014, system demonstrations}\ (\BPGS\ 55--60).
\newblock
\APACaddressPublisher{}{The Association for Computer Linguistics}.
\PrintBackRefs{\CurrentBib}

\bibitem [\protect \citeauthoryear {%
Maqbool%
\ \protect \BOthers {.}}{%
Maqbool%
\ \protect \BOthers {.}}{%
{\protect \APACyear {2018}}%
}]{%
Maqbool18}
\APACinsertmetastar {%
Maqbool18}%
\begin{APACrefauthors}%
Maqbool, B.%
, Azam, F.%
, Anwar, M\BPBI W.%
, Butt, W\BPBI H.%
, Zeb, J.%
, Zafar, I.%
\BDBL {}Umair, Z.%
\end{APACrefauthors}%
\unskip\
\newblock
\APACrefYearMonthDay{2018}{}{}.
\newblock
{\BBOQ}\APACrefatitle {A Comprehensive Investigation of {BPMN} Models Generation from Textual Requirements - Techniques, Tools and Trends} {A comprehensive investigation of {BPMN} models generation from textual requirements - techniques, tools and trends}.{\BBCQ}
\newblock
\BIn{} \APACrefbtitle {Information Science and Applications - {ICISA} 2018} {Information science and applications - {ICISA} 2018}\ (\BVOL~514, \BPGS\ 543--557).
\newblock
\APACaddressPublisher{}{Springer}.
\PrintBackRefs{\CurrentBib}

\bibitem [\protect \citeauthoryear {%
Mendling%
, Baesens%
, Bernstein%
\BCBL {}\ \BBA {} Fellmann%
}{%
Mendling%
\ \protect \BOthers {.}}{%
{\protect \APACyear {2017}}%
}]{%
DBLP:journals/dss/MendlingBBF17}
\APACinsertmetastar {%
DBLP:journals/dss/MendlingBBF17}%
\begin{APACrefauthors}%
Mendling, J.%
, Baesens, B.%
, Bernstein, A.%
\BCBL {}\ \BBA {} Fellmann, M.%
\end{APACrefauthors}%
\unskip\
\newblock
\APACrefYearMonthDay{2017}{}{}.
\newblock
{\BBOQ}\APACrefatitle {Challenges of smart business process management: An introduction to the special issue} {Challenges of smart business process management: An introduction to the special issue}.{\BBCQ}
\newblock
\APACjournalVolNumPages{Decis. Support Syst.}{100}{}{1--5}.
\newblock
\begin{APACrefURL} \url{https://doi.org/10.1016/j.dss.2017.06.009} \end{APACrefURL}
\newblock
\begin{APACrefDOI} \doi{10.1016/j.dss.2017.06.009} \end{APACrefDOI}
\PrintBackRefs{\CurrentBib}

\bibitem [\protect \citeauthoryear {%
Mendling%
, Leopold%
\BCBL {}\ \BBA {} Pittke%
}{%
Mendling%
\ \protect \BOthers {.}}{%
{\protect \APACyear {2015}}%
}]{%
Mendling:2014aa}
\APACinsertmetastar {%
Mendling:2014aa}%
\begin{APACrefauthors}%
Mendling, J.%
, Leopold, H.%
\BCBL {}\ \BBA {} Pittke, F.%
\end{APACrefauthors}%
\unskip\
\newblock
\APACrefYearMonthDay{2015}{}{}.
\newblock
{\BBOQ}\APACrefatitle {{25 Challenges of Semantic Process Modeling}} {{25 Challenges of Semantic Process Modeling}}.{\BBCQ}
\newblock
\APACjournalVolNumPages{International Journal of Information Systems and Software Engineering for Big Companies (IJISEBC)}{1}{1}{78--94}.
\PrintBackRefs{\CurrentBib}

\bibitem [\protect \citeauthoryear {%
Mendling%
, Leopold%
, Thom%
\BCBL {}\ \BBA {} van~der Aa%
}{%
Mendling%
\ \protect \BOthers {.}}{%
{\protect \APACyear {2019}}%
}]{%
DBLP:conf/refsq/MendlingLTA19}
\APACinsertmetastar {%
DBLP:conf/refsq/MendlingLTA19}%
\begin{APACrefauthors}%
Mendling, J.%
, Leopold, H.%
, Thom, L\BPBI H.%
\BCBL {}\ \BBA {} van~der Aa, H.%
\end{APACrefauthors}%
\unskip\
\newblock
\APACrefYearMonthDay{2019}{}{}.
\newblock
{\BBOQ}\APACrefatitle {Natural Language Processing with Process Models {(NLP4RE} Report Paper)} {Natural language processing with process models {(NLP4RE} report paper)}.{\BBCQ}
\newblock
\BIn{} \APACrefbtitle {Joint Proceedings of {REFSQ-2019} Workshops, Doctoral Symposium, Live Studies Track, and Poster Track co-located with the 25th International Conference on Requirements Engineering: Foundation for Software Quality {(REFSQ} 2019), Essen, Germany, March 18th, 2019} {Joint proceedings of {REFSQ-2019} workshops, doctoral symposium, live studies track, and poster track co-located with the 25th international conference on requirements engineering: Foundation for software quality {(REFSQ} 2019), essen, germany, march 18th, 2019}\ (\BVOL\ 2376).
\newblock
\APACaddressPublisher{}{CEUR-WS.org}.
\newblock
\begin{APACrefURL} \url{https://ceur-ws.org/Vol-2376/NLP4RE19\_paper04.pdf} \end{APACrefURL}
\PrintBackRefs{\CurrentBib}

\bibitem [\protect \citeauthoryear {%
Miller%
, Beckwith%
, Fellbaum%
, Gross%
\BCBL {}\ \BBA {} Miller%
}{%
Miller%
\ \protect \BOthers {.}}{%
{\protect \APACyear {1990}}%
}]{%
wordnet}
\APACinsertmetastar {%
wordnet}%
\begin{APACrefauthors}%
Miller, G\BPBI A.%
, Beckwith, R.%
, Fellbaum, C.%
, Gross, D.%
\BCBL {}\ \BBA {} Miller, K.%
\end{APACrefauthors}%
\unskip\
\newblock
\APACrefYearMonthDay{1990}{}{}.
\newblock
{\BBOQ}\APACrefatitle {WordNet: An on-line lexical database} {Wordnet: An on-line lexical database}.{\BBCQ}
\newblock
\APACjournalVolNumPages{International Journal of Lexicography}{3}{}{235--244}.
\PrintBackRefs{\CurrentBib}

\bibitem [\protect \citeauthoryear {%
Qian%
\ \protect \BOthers {.}}{%
Qian%
\ \protect \BOthers {.}}{%
{\protect \APACyear {2020}}%
}]{%
DBLP:conf/caise/0003W0LLZLW20}
\APACinsertmetastar {%
DBLP:conf/caise/0003W0LLZLW20}%
\begin{APACrefauthors}%
Qian, C.%
, Wen, L.%
, Kumar, A.%
, Lin, L.%
, Lin, L.%
, Zong, Z.%
\BDBL {}Wang, J.%
\end{APACrefauthors}%
\unskip\
\newblock
\APACrefYearMonthDay{2020}{}{}.
\newblock
{\BBOQ}\APACrefatitle {An Approach for Process Model Extraction by Multi-grained Text Classification} {An approach for process model extraction by multi-grained text classification}.{\BBCQ}
\newblock
\BIn{} \APACrefbtitle {Advanced Information Systems Engineering - 32nd Int. Conference, {CAiSE} 2020, Proceedings} {Advanced information systems engineering - 32nd int. conference, {CAiSE} 2020, proceedings}\ (\BVOL\ 12127, \BPGS\ 268--282).
\newblock
\APACaddressPublisher{}{Springer}.
\PrintBackRefs{\CurrentBib}

\bibitem [\protect \citeauthoryear {%
Quishpi%
, Carmona%
\BCBL {}\ \BBA {} Padr{\'{o}}%
}{%
Quishpi%
\ \protect \BOthers {.}}{%
{\protect \APACyear {2020}}%
}]{%
DBLP:conf/bpm/QuishpiCP20}
\APACinsertmetastar {%
DBLP:conf/bpm/QuishpiCP20}%
\begin{APACrefauthors}%
Quishpi, L.%
, Carmona, J.%
\BCBL {}\ \BBA {} Padr{\'{o}}, L.%
\end{APACrefauthors}%
\unskip\
\newblock
\APACrefYearMonthDay{2020}{}{}.
\newblock
{\BBOQ}\APACrefatitle {Extracting Annotations from Textual Descriptions of Processes} {Extracting annotations from textual descriptions of processes}.{\BBCQ}
\newblock
\BIn{} \APACrefbtitle {Business Process Management - 18th International Conference, {BPM} 2020, Proceedings} {Business process management - 18th international conference, {BPM} 2020, proceedings}\ (\BVOL\ 12168, \BPGS\ 184--201).
\newblock
\APACaddressPublisher{}{Springer}.
\PrintBackRefs{\CurrentBib}

\bibitem [\protect \citeauthoryear {%
Riefer%
, Ternis%
\BCBL {}\ \BBA {} Thaler%
}{%
Riefer%
\ \protect \BOthers {.}}{%
{\protect \APACyear {2016}}%
}]{%
Riefer16}
\APACinsertmetastar {%
Riefer16}%
\begin{APACrefauthors}%
Riefer, M.%
, Ternis, S\BPBI F.%
\BCBL {}\ \BBA {} Thaler, T.%
\end{APACrefauthors}%
\unskip\
\newblock
\APACrefYearMonthDay{2016}{}{}.
\newblock
{\BBOQ}\APACrefatitle {Mining process models from natural language text: A state-of-the-art analysis} {Mining process models from natural language text: A state-of-the-art analysis}.{\BBCQ}
\newblock
\APACjournalVolNumPages{Multikonferenz Wirtschaftsinformatik (MKWI-16), March}{}{}{9--11}.
\PrintBackRefs{\CurrentBib}

\bibitem [\protect \citeauthoryear {%
Schuler%
\ \BBA {} Palmer%
}{%
Schuler%
\ \BBA {} Palmer%
}{%
{\protect \APACyear {2005}}%
}]{%
verbnet}
\APACinsertmetastar {%
verbnet}%
\begin{APACrefauthors}%
Schuler, K\BPBI K.%
\BCBT {}\ \BBA {} Palmer, M\BPBI S.%
\end{APACrefauthors}%
\unskip\
\newblock
\APACrefYear{2005}.
\unskip\
\newblock
\APACrefbtitle {Verbnet: A Broad-Coverage, Comprehensive Verb Lexicon} {Verbnet: A broad-coverage, comprehensive verb lexicon}\ \APACtypeAddressSchool {\BUPhD}{}{}.
\unskip\
\newblock
\APACaddressSchool {USA}{University of Pennsylvania}.
\unskip\
\newblock
\APACrefnote{AAI3179808}
\PrintBackRefs{\CurrentBib}

\bibitem [\protect \citeauthoryear {%
van~der Aa%
, Carmona%
, Leopold%
, Mendling%
\BCBL {}\ \BBA {} Padr{\'{o}}%
}{%
van~der Aa%
\ \protect \BOthers {.}}{%
{\protect \APACyear {2018}}%
}]{%
Aa18}
\APACinsertmetastar {%
Aa18}%
\begin{APACrefauthors}%
van~der Aa, H.%
, Carmona, J.%
, Leopold, H.%
, Mendling, J.%
\BCBL {}\ \BBA {} Padr{\'{o}}, L.%
\end{APACrefauthors}%
\unskip\
\newblock
\APACrefYearMonthDay{2018}{}{}.
\newblock
{\BBOQ}\APACrefatitle {Challenges and Opportunities of Applying Natural Language Processing in Business Process Management} {Challenges and opportunities of applying natural language processing in business process management}.{\BBCQ}
\newblock
\BIn{} \APACrefbtitle {Proceedings of the 27th International Conference on Computational Linguistics, {COLING} 2018.} {Proceedings of the 27th international conference on computational linguistics, {COLING} 2018.}\ (\BPGS\ 2791--2801).
\newblock
\APACaddressPublisher{}{Association for Computational Linguistics}.
\PrintBackRefs{\CurrentBib}

\bibitem [\protect \citeauthoryear {%
van~der Aa%
, {Di Ciccio}%
, Leopold%
\BCBL {}\ \BBA {} Reijers%
}{%
van~der Aa%
\ \protect \BOthers {.}}{%
{\protect \APACyear {2019}}%
}]{%
derAaDiCiccio19}
\APACinsertmetastar {%
derAaDiCiccio19}%
\begin{APACrefauthors}%
van~der Aa, H.%
, {Di Ciccio}, C.%
, Leopold, H.%
\BCBL {}\ \BBA {} Reijers, H\BPBI A.%
\end{APACrefauthors}%
\unskip\
\newblock
\APACrefYearMonthDay{2019}{}{}.
\newblock
{\BBOQ}\APACrefatitle {Extracting Declarative Process Models from Natural Language} {Extracting declarative process models from natural language}.{\BBCQ}
\newblock
\BIn{} \APACrefbtitle {{Advanced Information Systems Engineering - 31st International Conference, {CAiSE} 2019, Proceedings}} {{Advanced Information Systems Engineering - 31st International Conference, {CAiSE} 2019, Proceedings}}\ (\BVOL\ 11483, \BPGS\ 365--382).
\newblock
\APACaddressPublisher{}{Springer}.
\PrintBackRefs{\CurrentBib}

\end{thebibliography}

\end{document}